\documentclass[sigconf,screen]{acmart}

\AtBeginDocument{%
  }

\setcopyright{acmlicensed}
\copyrightyear{2018}
\acmYear{2018}
\acmDOI{XXXXXXX.XXXXXXX}

\acmConference[Conference acronym 'XX]{Make sure to enter the correct
  conference title from your rights confirmation emai}{June 03--05,
  2018}{Woodstock, NY}
\acmISBN{978-1-4503-XXXX-X/18/06}

\copyrightyear{2024}
\acmYear{2024}
\setcopyright{acmlicensed}\acmConference[MM '24]{Proceedings of the 32nd ACM International Conference on Multimedia}{October 28-November 1, 2024}{Melbourne, VIC, Australia}
\acmBooktitle{Proceedings of the 32nd ACM International Conference on Multimedia (MM '24), October 28-November 1, 2024, Melbourne, VIC, Australia}
\acmDOI{10.1145/3664647.3681463}
\acmISBN{979-8-4007-0686-8/24/10}
\usepackage{multirow}
\begin{document}

%%
%% The "title" command has an optional parameter,
%% allowing the author to define a "short title" to be used in page headers.
\title[SaRO-GS]{4D Gaussian Splatting with Scale-aware Residual Field and Adaptive Optimization for Real-time Rendering of Temporally Complex Dynamic Scenes}

%%
%% The "author" command and its associated commands are used to define
%% the authors and their affiliations.
%% Of note is the shared affiliation of the first two authors, and the
%% "authornote" and "authornotemark" commands
%% used to denote shared contribution to the research.
\author{Jinbo Yan}
% \authornote{Both authors contributed equally to this research.}
\email{yjb@stu.pku.edu.cn}
% \orcid{1234-5678-9012}
% \author{G.K.M. Tobin}
% \email{webmaster@marysville-ohio.com}
\affiliation{%
  \institution{School of Electronic and Computer Engineering, Peking University}
  \city{Shenzhen}
  \country{China}
}

\author{Rui Peng}
\email{ruipeng@stu.pku.edu.cn}
\affiliation{%
  \institution{School of Electronic and Computer Engineering, Peking University}
  \city{Shenzhen}
  \country{China}
\email{larst@affiliation.org}}

\author{Luyang Tang}
\email{tly926@stu.pku.edu.cn}
\affiliation{%
  \institution{School of Electronic and Computer Engineering, Peking University}
  \city{Shenzhen}
  \country{China}
}

\author{Ronggang Wang}
\email{rgwang@pkusz.edu.cn}
% \authornotemark[1]
\authornote{Corresponding author.}
\affiliation{%
  \institution{School of Electronic and Computer Engineering, Peking University}
  \city{Shenzhen}
  \country{China}
}
\affiliation{%
  \institution{Pengcheng Laboratory}
  \city{Shenzhen}
  \country{China}
}
% \author{Huifen Chan}
% \affiliation{%
%   \institution{Tsinghua University}
%   \city{Haidian Qu}
%   \state{Beijing Shi}
%   \country{China}}

% \author{Charles Palmer}
% \affiliation{%
%   \institution{Palmer Research Laboratories}
%   \city{San Antonio}
%   \state{Texas}
%   \country{USA}}
% \email{cpalmer@prl.com}

% \author{John Smith}
% \affiliation{%
%   \institution{The Th{\o}rv{\"a}ld Group}
%   \city{Hekla}
%   \country{Iceland}}
% \email{jsmith@affiliation.org}

% \author{Julius P. Kumquat}
% \affiliation{%
%   \institution{The Kumquat Consortium}
%   \city{New York}
%   \country{USA}}
% \email{jpkumquat@consortium.net}

%%
%% By default, the full list of authors will be used in the page
%% headers. Often, this list is too long, and will overlap
%% other information printed in the page headers. This command allows
%% the author to define a more concise list
%% of authors' names for this purpose.

% \renewcommand{\shortauthors}{Yan et al.}

\begin{abstract}
Reconstructing dynamic scenes from video sequences is a highly promising task in the multimedia domain. While previous methods have made progress, they often struggle with slow rendering and managing temporal complexities such as significant motion and object appearance/disappearance. In this paper, we propose SaRO-GS as a novel dynamic scene representation capable of achieving real-time rendering while effectively handling temporal complexities in dynamic scenes. To address the issue of slow rendering speed, we adopt a Gaussian primitive-based representation and optimize the Gaussians in 4D space, which facilitates real-time rendering with the assistance of 3D Gaussian Splatting. Additionally, to handle temporally complex dynamic scenes, we introduce a Scale-aware Residual Field. This field considers the size information of each Gaussian primitive while encoding its residual feature and aligns with the self-splitting behavior of Gaussian primitives. Furthermore, we propose an Adaptive Optimization Schedule, which assigns different optimization strategies to Gaussian primitives based on their distinct temporal properties, thereby expediting the reconstruction of dynamic regions. Through evaluations on monocular and multi-view datasets, our method has demonstrated state-of-the-art performance. Please see our project page at \href{https://yjb6.github.io/SaRO-GS.github.io/}{https://yjb6.github.io/SaRO-GS.github.io/}.

\end{abstract}
%%
%% The code below is generated by the tool at http://dl.acm.org/ccs.cfm.
%% Please copy and paste the code instead of the example below.
%%


% \begin{CCSXML}
% <ccs2012>
%  <concept>
%   <concept_id>00000000.0000000.0000000</concept_id>
%   <concept_desc>Do Not Use This Code, Generate the Correct Terms for Your Paper</concept_desc>
%   <concept_significance>500</concept_significance>
%  </concept>
%  <concept>
%   <concept_id>00000000.00000000.00000000</concept_id>
%   <concept_desc>Do Not Use This Code, Generate the Correct Terms for Your Paper</concept_desc>
%   <concept_significance>300</concept_significance>
%  </concept>
%  <concept>
%   <concept_id>00000000.00000000.00000000</concept_id>
%   <concept_desc>Do Not Use This Code, Generate the Correct Terms for Your Paper</concept_desc>
%   <concept_significance>100</concept_significance>
%  </concept>
%  <concept>
%   <concept_id>00000000.00000000.00000000</concept_id>
%   <concept_desc>Do Not Use This Code, Generate the Correct Terms for Your Paper</concept_desc>
%   <concept_significance>100</concept_significance>
%  </concept>
% </ccs2012>
% \end{CCSXML}

% \ccsdesc[500]{Do Not Use This Code~Generate the Correct Terms for Your Paper}
% \ccsdesc[300]{Do Not Use This Code~Generate the Correct Terms for Your Paper}
% \ccsdesc{Do Not Use This Code~Generate the Correct Terms for Your Paper}
% \ccsdesc[100]{Do Not Use This Code~Generate the Correct Terms for Your Paper}

%%
%% Keywords. The author(s) should pick words that accurately describe
%% the work being presented. Separate the keywords with commas.

\begin{teaserfigure}
    \includegraphics[width=\textwidth]{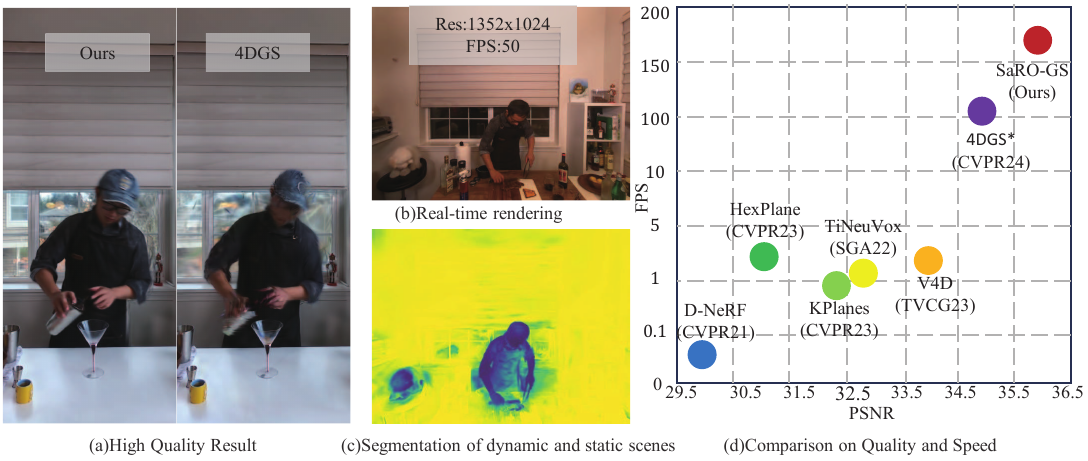}
    \caption{ Performance comparison with previous SOTA\cite{dnerf,hexplane,kplanes,v4d,tineuvox,4dgssplatting}. Our approach achieves higher-quality reconstruction in temporally complex scenes (a) while maintaining real-time rendering (b), with a certain improvement in performance(c). Additionally, we achieve dynamic scene segmentation without any prior information (c).*:measured by us at 400x400 resolution.
    }
    \label{fig:first}
    \Description{ Fully described in the text.}
\end{teaserfigure}

%%
%% This command processes the author and affiliation and title
%% information and builds the first part of the formatted document.
\maketitle
% \vspace{-0.5cm}
\section{INTRODUCTION}

% 动态场景的重建是沉浸影像的核心技术，支持着许多多媒体技术的发展，如：VR，AR，子弹时间，元宇宙，自由视角直播，视频会议，数字人，电影制作等。它可以从任意时间，任意位置和视角去对动态场景做渲染，这对提升用户对多媒体产品的体验至关重要。However，我们的目标是从一段时域上离散的视频序列中重建出一个连续的4D空间，这是一件十分棘手的事情，特别是在单目场景下。重建的质量是这个技术广泛应用的瓶颈，因为我们要accurately reconstruct the scene in spatial dimensions, but we also need to model its temporal variations。此外，用户对多媒体产品的实时性要求很高，实现实时渲染对提升用户的体验十分重要。现有的方法都难以做到同时兼顾高质量的重建和实时渲染，这正是我们这篇工作所要解决的问题。

The reconstruction of dynamic scenes is pivotal for immersive imaging, driving advancements in various multimedia technologies such as VR, AR, and metaverse. Our objective is to reconstruct a continuous 4D space from a discrete temporal video sequence. However, this endeavor faces several challenges. Firstly, the reconstruction quality acts as a bottleneck for widespread adoption, requiring accurate capture of spatial dimensions and temporal variations in dynamic scenes. Additionally, there's a growing demand for real-time interaction in multimedia products to boost user engagement, highlighting the importance of achieving real-time rendering. Nevertheless, existing methods struggle to achieve both high-quality reconstruction and real-time rendering simultaneously, precisely the issue our approach aims to tackle.
% Reconstructing dynamic scenes from a video sequence and enabling view synthesis from any perspective at any time is a critical challenge in the multimedia domain, with significant implications for various fields such as virtual reality, augmented reality, video generation, autonomous driving, robotics, and others. 
% However, this problem becomes highly challenging because not only do we need to accurately reconstruct the scene in spatial dimensions, but we also need to model its temporal variations. Choosing a suitable modality to represent this 4D scene and optimizing the continuous 4D space through discrete sampling from input video sequences is crucial for addressing this issue.

% Our goal is to achieve a representation of dynamic scenes that can establish a 4D space from monocular or multi-view video input sequences, supporting view synthesis from any perspective at any time. Additionally, it should enable real-time rendering and modeling of complex temporal information.
Recent advancements in dynamic scene reconstruction have been achieved through methods based on NeRF \cite{nerf} and 3DGS \cite{3dgs}. NeRF employs an implicit field to model static scenes and achieves photo-realistic view synthesis. Many extensions of NeRF to dynamic scenes either utilize deformation fields and canonical fields to model the motion of objects relative to canonical frames over time\cite{dnerf,hypernerf,devrf,neuraltraj,flowforwardnerf,nerfplayer}, or decompose the 4D volume into spatial-only and spatial-temporal spaces\cite{kplanes,hexplane,tensor4d,im4d}, representing space through combinations of dimensionally reduced features. While significant progress has been made in rendering quality, these methods face a significant disadvantage in rendering speed. The emergence of 3DGS has enabled real-time rendering of dynamic scenes. Some methods \cite{4dgssplatting,realtime4dgs,spacetimegs,deformable3dgs,scgs} have attempted dynamic scene modeling based on 3DGS. However, they either struggle to model temporally complex scenes such as object appearances and disappearances \cite{4dgssplatting,deformable3dgs,scgs} or overlook the spatiotemporal information in the scene \cite{realtime4dgs,spacetimegs}, resulting in disadvantages when dealing with temporally complex dynamic scenes.

To address the aforementioned challenges, we propose SaRO-GS, aiming to achieve real-time rendering while maintaining high-quality reconstruction of temporally complex dynamic scenes. SaRO-GS comprises a set of Gaussian primitives in 4D space and a Scale-aware Residual Field. Each Gaussian receives a unique optimization schedule based on its distinct temporal properties through an Adaptive Optimization strategy.
To address the issue of slow rendering speeds, Gaussian primitives in 4D space can be projected to 3D based on their temporal properties and residual features obtained from the Scale-aware Residual Field. Then we can achieve real-time rendering leveraging the fast differentiable rasterizer introduced by 3DGS.
% for real-time rendering of dynamic scenes
For high-quality modeling of temporally complex scenes, we employ the following strategies: Firstly, each 4D Gaussian primitive possesses temporal properties, including temporal position and lifespan. The lifespan allows us to model the appearance and disappearance of objects in dynamic scenes, while the temporal position of Gaussians spans the entire temporal range, rather than being fixed at frame 0 as in previous methods. 
% With these temporal properties, we can also achieve dynamic scene segmentation without any prior information, as discussed in Sec. 8.
% Unlike previous deformation-based methods, 
% Momentum-driven 4D splatting is based on the temporal position of each point rather than being anchored to the first frame, providing greater flexibility. Our representation also allows us to achieve dynamic scene segmentation without any prior information, as shown in Fig. \ref{fig:first}(c).
Additionally, we incorporate scale information of Gaussian primitives into the Residual Field to accommodate their ellipsoidal nature. By encoding the region that the Gaussian primitives occupy rather than just their position, we ensure accurate feature extraction and align with the self-splitting behavior of Gaussian primitives. Thirdly, we introduce an Adaptive Optimization strategy, where unique optimization strategies are assigned to each Gaussian primitive based on its temporal properties, facilitating faster reconstruction of dynamic regions.

% integral-based optimization schedule, wherein unique optimization strategies are assigned to each Gaussian primitive based on its sampling probability during our Momentum-driven 4D splatting process, thereby facilitating faster reconstruction of dynamic regions.

We extensively evaluated our approach on monocular and multi-view dynamic scene datasets, comprising both real and synthetic scenes. Both quantitative and qualitative results demonstrate that our method achieves high-quality rendering in real time and effectively handles temporal complexities in dynamic scenes. Our contributions are summarized below.
% \vspace{-2ex}

\begin{itemize}
    % \item We propose a Mos representation to model dynamic scene by leveraging 4D Gaussian primitives and a momentum field . Combining with our Momentum-driven 4D splatting process, we achieve real-time rendering and  capture fine-grained temporal information.
    \item We propose a Scale-aware Residual Field, incorporating the scale information of Gaussians. This results in a more precise spatiotemporal representation, considering Gaussian primitives' ellipsoidal nature and self-splitting behavior.
    \item We introduce an Adaptive Optimization strategy, assigning unique optimization schedule to Gaussians based on their unique temporal properties, enhancing the reconstruction of dynamic areas.
    \item Our SaRO-GS excels in managing temporally complex scenarios, delivering state-of-the-art performance in both the reconstruction quality and rendering speed. It achieves an 80x rendering speed improvement compared to NeRF-based methods as shown in Fig. \ref{fig:first}. SaRO-GS is versatile, applicable to both monocular and multi-view scenarios and can also achieve dynamic scene segmentation without any prior.
    % \item We introduce a per-Gaussian optimization schedule to address the uneven sampling in 4D space during our Momentum-driven 4D splatting process, enhancing the reconstruction of dynamic areas.
\end{itemize}

\vspace{-1ex}
\section{RELATED WORK}
% We first review methods for static scene reconstruction based on Neural Rendering. Then, we delve into dynamic scene reconstruction, discussing approaches based on NeRF and 3DGS separately.
% \vspace{-0.7cm}
\subsection{Static Scene Representation }
Significant progress has been made in various fields related to the task of static scene representation, including image matching\cite{lowe2004distinctive,sarlin2020superglue,shen2024gim}, camera calibration\cite{schoenberger2016mvs,schoenberger2016sfm}, depth estimation\cite{eigen2014depth,garg2016unsupervised,gao2021pdanet,peng2022rethinking}, and Multi-View Stereo\cite{yao2018mvsnet,yao2019recurrent,peng2022rethinking,zhang2023bi,zhang2023n2mvsnet,zhang2023geomvsnet,xiong2023cl}, among others. Recently, there has been a growing interest in the use of Neural Radiance Fields as a means of representing static scenes. NeRF\cite{nerf}, as a representative work, models static scenes as a radiance field and synthesizing images through volume rendering. 
The photo-realistic view synthesis capability of NeRF has inspired a series of works across various domains, including enhancing rendering quality\cite{mipnerf21,nerfsr,depth-nerf,mirror-nerf,nerf++,zipnerf,nerfbr,mipnerf360}, sparse inputs\cite{wahnerf,regnerf,xu2024hdpnerf}, surface representation and segmentation\cite{wang2021neus,peng2023gens,zheng2024surface}, accelerating training and rendering\cite{instantngp,snerg,plenxoel,dvgo,plenoctree,tensor4d,merf,trimip,dignerf,Kilonerf,zipnerf}, as well as human modeling\cite{humannerf,neuralbody,anihuman,monohuman,zheng2024pku}, among others.
% 对神经场景重建的加速训练和渲染也是我们所关注的。NeRF由于基于纯隐式的方法，一个像素的渲染就需要密集的光线采样以及MLP推理，使其速度过慢。有许多方法采取了显式的结构来对其加速，将特征存储在显式的网格或hash table中，这样可以使他们采取更小的MLP，有的方法采用了SH系数来表示view-dependent，甚至可以做到不需要MLP。但这些方法仍然无法拜托渲染时需要密集光线采点的限制，无法较好的跳过空白区域。
% Accelerating training and rendering for neural scene reconstruction is also a concern for us. Due to its purely implicit nature, NeRF\cite{nerf} requires dense ray sampling and MLP inference for rendering a single pixel, resulting in slow processing speeds. Various methods have been devised to accelerate NeRF by introducing explicit structures\cite{instantngp,dvgo,plenxoel,tensor4d,snerg,merf,trimip,plenoctree}, such as storing features in explicit grids or hash tables. This allows for the adoption of smaller MLPs. Plenoxels\cite{plenxoel} use spherical harmonics (SH) coefficients to represent view-dependent effects, even eliminating the need for MLPs entirely. However, these methods still struggle to overcome the limitation of dense ray sampling during rendering and fail to efficiently skip empty regions.
Recently, there has been a breakthrough in high-quality view synthesis and real-time rendering with 3D Gaussian Splatting\cite{3dgs} and relate works\cite{mipsplatting,fsgs,gps-gaussian,dnggaussian}, garnering significant attention in the scene reconstruction field.
% It represents scenes using 3D Gaussian primitives and adopts a point-based rendering approach. 
% Some efforts are dedicated to improving their rendering quality and in applications like AIGC, garnering significant attention in the scene reconstruction field.
% It represents scenes using 3D Gaussian primitives and adopts a point-based rendering approach. 
% Some efforts are dedicated to improving their rendering quality, garnering significant attention in the scene reconstruction field.
% Some efforts are dedicated to improving their generalization, rendering quality, and reconstruction capability under sparse viewpoints. This has also spurred some work in applications such as SLAM and AIGC, garnering significant attention in the scene reconstruction field.
 
% However, extending this technique to dynamic scenes presents a new and challenging problem, 一些方法做了尝试但难以达到在时域复杂场景下的理想效果，which is what our SRH-GS aims to address.
\vspace{-0.4cm}
\subsection{Dynamic Scene Representation}
% 简单的为NeRF增加一个时间域并不能达到理想的效果，一些工作采取了基于静态场和变形场的方法来表示动态场景[D-NeRF,NeRF-player,ReRF]。一般的做法是选取
% 目前的基于NeRF的动态场景表示方法主要可以分为三类
Expanding static scene representation to dynamic scenes is not a simple task. 
% Representing the entire dynamic scene by directly building static scenes for each frame loses the correlation between time and space. This results in significant computational and storage costs and fails to achieve view synthesis at arbitrary times. 
Some NeRF based methods\cite{dnerf,hypernerf,devrf,neuraltraj,flowforwardnerf,nerfplayer} have made progress based on deformation fields, modeling the entire scene as a canonical field and a deformation field. They use the deformation field to represent the association between sampled points under different frames and the static canonical. 
% However, these methods struggle to handle complex temporal scenarios, such as object appearance and disappearance. 
% NeRFPlayer\cite{nerfplayer} addresses the appearance of new objects by employing a sliding window to represent temporal features but overlooks spatiotemporal correlations, leading to a linear increase in storage cost with the number of video frames. 
Other methods\cite{kplanes,hexplane,tensor4d,im4d,maskedspacetimehash} reduce the dimensionality of the 4D space by decomposing it into a set of planar grids or hash grids. This approach effectively models temporal correlations through spatiotemporal grids.
Furthermore, some methods\cite{rerf,streamrf} adopt a streaming strategy to model residuals between adjacent frames, which is suitable for real-time transmission and decoding.
However, these rendering approachs based on NeRF requires dense sampling along rays during rendering, limiting the possibility of real-time rendering.
% but there is still significant room for improvement in terms of quality.
% 一些方法基于变形场取得了一些进展，他们将整个场景建模为一个静态场和变形场，通过变形场表示不同帧下采样点和静态场的关联，但这些方法难以处理时域上的复杂情形，如物体的出现和消失.NeRFPlayer针对新物体的出现进行了处理，采用滑动窗口来表示时域的特征，但忽视了时空相关性，存储代价随视频帧数线性增长.另一些方法对4D空间进行降维，将4D空间降解为一组平面网格，将4D的采样点向平面投影得到对应的特征，这种方法能通过spatio-temporal plane很好的建模时域相关性.

% 0.直接把时域输入MLP中
% N3D,
% 1.静态场变形场
% 把4d采样点warp到静态场：hypernerf,d-nerf,DeVRF，nerfplayer
% 基于轨迹的方法：Neural Trajectory Fields
% 静态场warp到其他帧：Forward Flow for Novel View Synthesis of Dynamic Scenes，
% 2.空间域时间域分解
% hexplane,kplane,tensor4d,im4d
% 3.逐帧去建
% Streaming Radiance Fields for 3D Video Synthesis
% rerf
% 每一帧直接建：ENeRF
% \vspace{-0.45cm}
% \subsection{Dynamic Scene Representation Based on 3D Gaussian}
% 有一些同期的基于3D Gaussian表示动态场景的工作，Dynamic Gaussian和3dstream采取了online训练的策略，让Gaussian primitive随着帧增加逐渐evolve。4dgs采用了一个Hexplane来建模Gaussian primitive随着采样时间的变化，但难以建模时域上的复杂情形，如物体的出现和消失. iclr4dgs为Gaussian增加了时间的维度，可以让4D Gaussian decomposed into a conditional 3D Gaussian and a marginal 1D Gaussian，与之相比，我们通过一个SRH field来建模Gaussian primitive从4D向3D投影的残差，采用显隐式混合的方式，更能结合时空相关性。还有一些方法如spacetime gs不适用于多目情况，deformable gs在多目情况下效果会下降很多，而我们的方法通过实验验证，在单目和多目的情况下都能取得不错的效果。

There are some concurrent works based on 3D Gaussian for dynamic scenes. \cite{dynamic3dgs} uses an online strategy to model dynamic scenes frame by frame, \cite{4dgssplatting} uses a Hex-planes to model the changes of Gaussian primitives over sampling time. Both of them struggle to handle temporal complexities, such as  significant motion and object appearances/disappearances. \cite{realtime4dgs} introduces a time dimension to Gaussian, enabling 4D Gaussian to be decomposed into a conditional 3D Gaussian and a marginal 1D Gaussian. In comparison, we propose an Scale-aware Residual field to model the residual of Gaussian primitives projected from 4D to 3D, employing an explicit-implicit blending approach that better incorporates spatiotemporal correlations. Other methods like \cite{spacetimegs} are not suitable for monocular scenarios, and \cite{deformable3dgs} cannot be utilized in multi-view scenes. In contrast, our approach, validated through experiments, demonstrates promising results in both single-view and multi-view scenarios.

% Dynamic Gaussian and 3dstream adopt an online training strategy, allowing Gaussian primitives to evolve gradually as frames increase.
% \subsection{Template Styles}

% \subsection{Template Parameters}

% In addition to specifying the {\itshape template style} to be used in
% formatting your work, there are a number of {\itshape template parameters}
% which modify some part of the applied template style. A complete list
% of these parameters can be found in the {\itshape \LaTeX\ User's Guide.}

% Frequently-used parameters, or combinations of parameters, include:
% \begin{itemize}
% \item {\verb|anonymous,review|}: Suitable for a ``dual-anonymous''
%   conference submission. Anonymizes the work and includes line
%   numbers. Use with the \verb|\acmSubmissionID| command to print the
%   submission's unique ID on each page of the work.
% \item{\verb|authorversion|}: Produces a version of the work suitable
%   for posting by the author.
% \item{\verb|screen|}: Produces colored hyperlinks.
% \end{itemize}

% This document uses the following string as the first command in the
% source file:
% \begin{verbatim}
% \documentclass[sigconf,authordraft]{acmart}
% \end{verbatim}

\section{PRELIMINARY}
\subsection{3D Gaussian Splatting}
Given a set of input images of a static scene along with their corresponding camera parameters, 3D Gaussian Splatting (3DGS) initiates the reconstruction of the static scene from an initial point cloud, employing 3D Gaussians as primitives. This approach enables high-quality real-time novel view synthesis.

In 3DGS, each Gaussian primitive encompasses a set of attributes, including 3D position $\mu_{3d}$, opacity $\alpha$, and covariance matrix $\Sigma$. A 3D Gaussian $\mathcal{G}$ can be represented as:
\begin{equation}
    G(x) = e^{-\frac{1}{2}(x-\mu)^T\Sigma^{-1}(x-\mu)}
    \label{eq:gx}
\end{equation}
For optimization convenience, 3DGS employs a scaling matrix $S$ and a rotation matrix $R$ to represent covariance, stored as a 3D vector $s$ for scaling and a quaternion $q$ for rotation. 
\begin{equation}
    \Sigma = RSS^TR^T
\end{equation}
Additionally, 3DGS utilizes SH coefficients to represent view-dependent color.

Based on the fast differentiable rasterizer implemented by 3DGS, we can achieve rapid image rendering through Gaussian Splatting. To get the rendering images from a given viewpoint, we should first project 3D Gaussian primitives to 2D. Specifically, for a given viewpoint transformation matrix $W$ and a projection matrix $K$, we can obtain the covariance and the position in 2D space:
\begin{equation}
\begin{aligned}
      \Sigma^{2D} &= (JW\Sigma W^T J^T)_{1:2,1:2}\\
    \mu^{2D} &= {K( \frac{W\mu}{(W\mu)_z})}_{1:2}  
\end{aligned}
    \label{eq:mu2d_cov2d}
\end{equation}
where $J$ is the Jacobian of the affine approximation of the projetive transformation. And we can get the 2D Gaussian $\mathcal{G}^{2d}$ based on Eq. \ref{eq:gx}.

After sorting the Gaussian primitives in 2D space based on depth, we can obtain the color of the specified pixel in the image:
\begin{equation}
    c(x) = \sum_{i=1}^N c_i \alpha_i\mathcal{G}_i^{2D} \prod_{j=1}^{i-1}(1-\alpha_j \mathcal{G}_j^{2D}(x))
    \label{eq:alphablending}
\end{equation}
Here, $c_i$ represents the view-dependent color obtained by combining the SH coefficients of $\mathcal{G}_i$ with the viewing direction.

% 3DGS directly optimizes the attributes of Gaussian primitives based on the loss between the rendered image and the original image. Additionally, every 100 iterations, it densifies primitives based on view-space positional gradients that exceed a certain threshold, aiming to achieve higher precision in scene reconstruction.
\subsection{4D Volume Representation Based on Hex-planes}
\label{sec:hexplane}
Previous works modeling dynamic scenes using plane field encoders mostly employed hexplanes $P$, which encompass spatial-only planes $P_{so} =\{ P_{x,y},P_{y,z},P_{x,z} \}$ and spatiotemporal planes $P_{st} =\{ P_{x,t},P_{y,t},P_{z,t} \}$. Each plane $P_{i,j}$ in $P$ is a $M \times N \times N$ two-dimensional grid, where $M$ represents the feature dimension and $N$ represents the spatiotemporal resolution of the grid. For a 4D sample point $q$ = (x,y,z,t), we perform interpolation based on its projected coordinates on the six grids to obtain the corresponding feature for this point:
\begin{equation}
\begin{aligned}
    f(q)& = \prod_{i,j \in C} \psi_{bi}(\pi_{i,j}(q);P_{i,j}) \\
C = \{(x,y)&,(x,z),(x,t),(y,z),(y,t),(z,t)\}
\end{aligned}
\end{equation}
Here, $f(q)$ is an $M$-dimensional feature, $\psi_{bi}$ represents bilinear interpolation, and $\pi_{i,j}(q)$ denotes the projected coordinates of sample point $q$ on $P_{i,j}$.

\begin{figure*}[h]
  \centering
    % \fbox{\rule{0pt}{2.5in} \rule{0.9\linewidth}{0pt}}
    \setlength{\abovecaptionskip}{0.cm}

  \includegraphics[width=0.8\linewidth,height=0.4\textwidth]{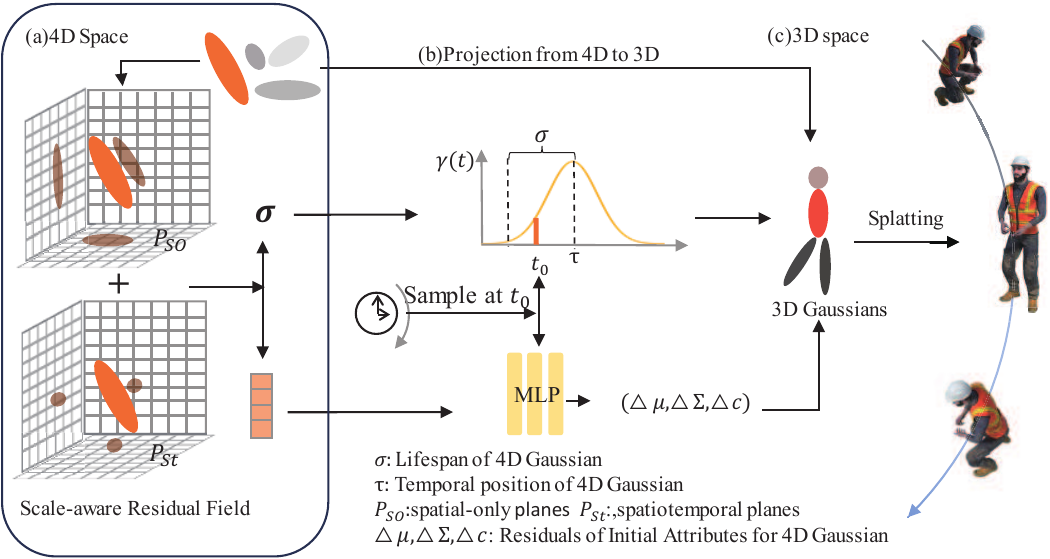}
  \caption{
  The overall pipeline of SaRO-GS.
  % (a)在4D空间中，我们同时优化了一组4D Gaussian primitives和一个scale-aware Momentum field. Gaussian primitive in 4D space包含的属性有一个4D position和一些初始属性：a,b,c.Momentum field用Hex-Plane来表示4D的volume，由三个spatio-only planes和三个spatiotemporal planes组成。每个Gaussian primitive在可以通过Momentum field得到一个 momentum feature和一个lifespan.他们表示了Gaussian primitive在时域上的特性
  (a)In 4D space, we simultaneously optimize a set of 4D Gaussians and a scale-aware Residual Field $\mathcal{M}$. 
  % Each Gaussian has a 4D location $\mu^{4D}=(x,y,z,\tau)$, where $\tau$ represents the temporal position.
  % A Gaussian primitive in 4D space consists of a 4D position and some initial attributes: a, b, c. 
  % The Momentum field is represented by a Hex-Plane, which captures the 4D volume using three spatio-only planes and three spatiotemporal planes. 
  % Each Gaussian primitive, when combined with the Momentum field, yields a momentum feature and a lifespan. 并且在产生特征的过程中Momentum field会考虑Gaussian primitive的尺度信息.
  When combined with $\mathcal{M}$, each Gaussian generates a residual feature and a lifespan $\sigma$. 
  % Furthermore, \mathcal{M} takes into account the scale information of the Gaussian primitive during the feature generation process.
  They both represent the temporal characteristics of the Gaussian primitive.
  (b)Given a sampling time $t_0$, we can compute the survival status $\gamma(t_0)$ of the Gaussian 
  % We then multiply it by the initial opacity $\alpha^{4D}$ to obtain the opacity in 3D space $\alpha^{3D}$. 
 and decode the residual feature of the Gaussian at time $t_0$ using an MLP, yielding residual of atteibutes. 
  Finally, we combine these residuals with the initial attributes of the Gaussian in 4D space to get the 3D Gaussian representation.(c) Once we obtain the representation of the 3D Gaussian, we can generate rendered images using Gaussian Splatting.
  }
  \Description{ Fully described in the text.}
  \label{fig:fiiiig}
  \vspace{-0.5cm}
\end{figure*}

\section{METHOD}

In this section, we initiate with the presentation of 
% the representation method and 
overall pipeline of SaRO-GS in Sec. \ref{sec:pipeline}. Subsequently, we explore the Scale-aware Residual Field in Sec. \ref{sec:momentum_field}, with a particular emphasis on its consideration of scale information for Gaussian primitives. Following this, we elaborate on the Adaptive Optimization strategy employed in Sec. \ref{sec:optim}. Finally, we delineate the loss function and regularization terms utilized in our approach in Sec. \ref{sec:loss}.
\subsection{Representation of SaRO-GS}
\label{sec:pipeline}
To represent a dynamic scene, we utilize a set of Gaussian primitives $\mathcal{G}^{4D}$ in 4D space alongside Scale-aware Residual Field $\mathcal{M}$, as shown in Fig. \ref{fig:fiiiig}(a). Each 4D Gaussian primitive $\mathcal{G}_i^{4D}$ possesses a temporal position $\tau_i$,it is learned alongside its 3D position (x,y,z), forming a 4D location $\mu^{4D}=(x,y,z,\tau)$. Together with the initial attributes $\Sigma^{4D}$, $c^{4D}$ and $\alpha^{4D}$, a 4D Gaussian primitive and its residual feature $f$ can be represented as follows:
\begin{equation}
     \mathcal{G}^{4D} = (\mu^{4D},\Sigma^{4D},c^{4D},\alpha^{4D}),
\end{equation}
\begin{equation}
    f = \mathcal{M}(\mathcal{G}^{4D}). 
\end{equation}
 $\Sigma^{4D},c^{4D}$ and $\alpha^{4D}$ respectively represent the initial covariance, color, and opacity of the Gaussian primitive in 4D space. Similar to 3D Gaussian, we employ quaternion rotation $q^{4D}$ = ($q_a,q_b,q_c,q_d$) and scaling vectors $s^{4D}$= ($s_x,s_y,s_z$) to represent covariance, and utilize SH (Spherical Harmonics) coefficients to depict view-dependent color.

 To address complex temporal scenarios such as object appearance and disappearance, each Gaussian primitive should have a lifespan to indicate how long it can survive in the temporal domain.
 In order to effectively integrate the Scale-aware Residual Field $\mathcal{M}$ with our 4D Gaussian primitives and leverage the spatiotemporal characteristics of $\mathcal{M}$, we employ a tiny MLP $\mathcal{F}_w$ to perform inference on $f_i$ and compute the lifespan $\sigma_i$ of $\mathcal{G}_i^{4D}$:
\begin{equation}
    \sigma_i = \mathcal{F}_w(f_i) = \mathcal{F}_w(\mathcal{M}(\mathcal{G}_i^{4D})).
\label{eq:sigma}
\end{equation}
Therefore, in our 4D space, each Gaussian primitive $\mathcal{G}_i^{4D}$ can obtain a residual feature $f_i$ and a lifespan $\sigma_i$ through $\mathcal{M}$. $\{\mathcal{G}_i^{4D},f_i,\sigma_i\}$ fully represents both the initial attributes and temporal characteristics of a Gaussian primitive in 4D space.
% \noindent \textbf{Momentum-driven 4D Splatting:}

Once the sampling time $t_0$ are given, we need to project the Gaussian primitives from 4D space to 3D space, as shown in Fig. \ref{fig:fiiiig}(b). We first need to examine whether $\mathcal{G}_i^{4D}$ still survives at the current sampling time $t_0$. Inspired by \cite{spacetimegs}, we adopt a Gaussian-like state function $\gamma(t)$ to model the state of $\mathcal{G}_i^{4D}$ as it varies with the sampling time $t$:
\begin{equation}
\setlength{\abovedisplayskip}{3pt}
\setlength{\belowdisplayskip}{3pt}
    \gamma_i(t) = e^{-k{\frac{t-\tau_i}{\sigma_i}}^2}.
    \label{eq:gamamat}
\end{equation}
% 其中t表示当前的采样时间，a表示g的temporal position. In practice, k is set to 4. When the sampling time t 从g的temporal position 逐渐远离, l(t)会从1逐渐减小. and when the sampling time reaches the Gaussian lifespan，l(t) will decrease to 0.01,这表示g在t下近乎失活，即在投影到3D空间后，g应该是不可见的.我们可以以此来表示G在3D空间中的opacity:
Where $t$ represents the sampling time, and $\tau_i$ represents the temporal position of $\mathcal{G}_i^{4D}$. In practice, $k$ is set to 4. As the sampling time $t$ gradually moves away from the temporal position of $\mathcal{G}_i^{4D}$, $\gamma(t)$ decreases from 1. When the sampling time reaches the Gaussian lifespan $\sigma_i$, $\gamma(t)$ will decrease to 0.01, indicating that $\mathcal{G}_i^{4D}$ is nearly inactive at $t$, which means it should be invisible when projected into 3D space. So for a given sampling time $t_0$, We can utilize this state function to represent the opacity of $\mathcal{G}_i^{4D}$ in 3D space after projection:
\begin{equation}
    \alpha_i^{3D} = \alpha_i^{4D} \times \gamma_i(t_0)
    \label{eq:alpha3d}
\end{equation}

Apart from opacity, other features of $\mathcal{G}_i^{4D}$ also vary with sampling time $t$ when projected into 3D space. We can utilize a set of MLPs $\mathcal{F}_{\theta}$ to decode the residual feature $f_i$ of 
$\mathcal{G}_i^{4D}$ at the sampling time 
$t$, thereby obtaining the residual of the projected attributes as they vary with the sampling time $t$.
\begin{equation}
    \Delta \mu_i(t) ,\Delta \Sigma_i(t) , \Delta c_i(t) = \mathcal{F}_\theta (f(\mathcal{G}_i^{4D}) , t-\tau_i )
    \label{eq:delta}
\end{equation}
$\Delta \mathbf{\mu}_i(t) ,\Delta \mathbf{\Sigma}_i(t)$ and $\Delta c_i(t)$ respectively represent the residuals of position, covariance, and color. Here we decode using 
$t-\tau_i$ instead of just 
$t$, as we aim to obtain the residual relative to the initial attribute of 
$\mathcal{G}_i^{4D}$ in 4D space, where 
$\mathcal{G}_i^{4D}$ is temporally positioned using 
$\tau_i$.

Therefore , we can obtain the attributes of projected $\mathcal{G}_i^{3D}$ at a given time $t_0$:
\begin{equation}
    \mathbf{\mu}_i^{3D} = \mathbf{\mu}_i^{4D}[:3] + \Delta \mathbf{\mu}_i(t_0),
    \label{eq:plus_mu_attributes}
\end{equation}
\begin{equation}
    \Sigma_i^{3D} = {\Sigma}_i^{4D}+ \Delta \mathbf{\Sigma}_i(t_0),
    \label{eq:plus_cov_attributes}
\end{equation}
\begin{equation}
    c_i^{3D} = c_i^{4D} + \Delta c_i(t_0).
    \label{eq:plus_color_attributes}
\end{equation}
$\mathbf{\mu}_i^{4D}[:3]$ represents extracting the xyz components of $\mathbf{\mu}_i^{4D}$ as the initial 3D location, and
$\Delta \mathbf{\Sigma}_i$ can be decomposed into 
$\Delta \mathbf{\Sigma}_i[:3]$ and 
$\Delta \mathbf{\Sigma}_i[3:]$, representing the residuals of the three-dimensional scaling vector and quaternion rotation about 
$\mathcal{G}_i^{4D}$, respectively. The changes to ${\Sigma}_i^{4D}$ and $c_i^{4D}$ are achieved by adjusting the corresponding quaternion rotation, scaling vectors, and SH coefficients.

Therefore, according to Eq. [\ref{eq:sigma}-\ref{eq:plus_color_attributes}], Gaussian primitives in 4D space evolve within their lifespan as the sampling time varies, projecting into 3D space at a given sampling time. Then, based on Eq. \ref{eq:mu2d_cov2d}\ref{eq:alphablending}, we employ 3DGS to render 3D Gaussians, obtaining rendered images from a given camera viewpoint, as shown in Fig. \ref{fig:fiiiig}(c).

\subsection{Scale-aware Residual Field}
\label{sec:momentum_field}
% To fully integrate the spatiotemporal information of the scene and save computational resources, we adopt hexplanes to represent our 4D Momentum Field, which is 由spatial only planes 和spatiotemporal planes 组成，如果section 1所描述的那样. However, disregarding the size of Gaussian primitives and only use the 4D position 去向平面投影去特征 would
% result in 取到错误的momentum feature.

\begin{figure}[h]
  \centering
  \setlength{\abovecaptionskip}{0.cm}

    % \fbox{\rule{0pt}{2.5in} \rule{0.9\linewidth}{0pt}}
  \includegraphics[width=0.8\linewidth]{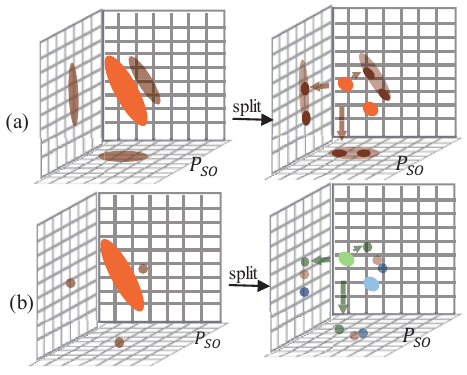}

  \caption{The impact of scale is not taken into account in Gaussian self-splitting. (a)When size information is considered, the features of the split Gaussian remain similar to its parent Gaussian. (b)Otherwise, the split Gaussian will have features different from its parent Gaussian}
\label{fig:plane}
\vspace{-0.6cm}
\end{figure}

To fully integrate the spatiotemporal information of the scene and save computational resources, we adopt hexplanes to represent our Scale-aware Residual Field $\mathcal{M}$, which consists of spatial-only planes and spatiotemporal planes, as described in Sec. \ref{sec:hexplane}. 

However, neglecting the size of Gaussian primitives and solely projecting them onto planes based on their 4D positions for feature extraction would lead to incorrect residual features.
First, Gaussian primitives can be approximated as ellipsoids. Therefore, when projecting a Gaussian primitive onto a spatial-only plane, we obtain an elliptical region instead of a single point, as in the current NeRF-based approach. Thus, the corresponding feature for a Gaussian primitive should be a combination of all the areas it occupies in the plane. Secondly, if we follow the self-splitting strategy of 3DGS and split a large Gaussian primitive into smaller ones, they would have different residual features, significantly deviating from those of their parent primitive, which contradicts our original intention, as shown in Fig. \ref{fig:plane}. So, finding an appropriate method to encode the region projected by Gaussian is crucial.

We propose a scale-aware Residual Field to address the aforementioned issue, which decompose the 4D space into three spatial-only planes $P_{so}$ and three spatiotemporal planes $P_{st}$. Given that the size of Gaussian primitives impacts their projection only within the spatial-only plane $P_{so}$, we specifically consider only employing scale-aware encoding within them, as shown in Fig. \ref{fig:fiiiig}.

For every spatial-only planes $P_{i,j}$, we employ a MipMap stack to represent features at different spatial scales in the scene. The level 0 of the MipMap stack $P_{i,j}^0$ is a feature map with shape  $M \times N \times N$ ,which has the smallest spatial scale $\ddot s^{0}$ among all levels. And remaining levels $P_{i,j}^l$ in the Mipmap stack are obtained by computing thumbnails based on the features of the previous level, where the width and height are reduced by a factor of 2 each. Taking $P_{x,y}$ as an example, the relationship between their spatial scale is as follows:
\begin{equation}
\begin{aligned}
       \ddot s_x^{0} = \frac{\mathcal{B}_{max}^x - \mathcal{B}_{min}^x}{N} \quad\quad&\quad\quad \ddot s_y^{0} = \frac{\mathcal{B}_{max}^y - \mathcal{B}_{min}^y}{N}\\
       \ddot s^{l+1} = 2 \times \ddot s^{l} \quad\quad\quad&\quad\quad\quad l \in [0,L-1]
% \quad\quad\quad \ddot s_0^{y} = \frac{\mathcal{B}_{max}^y - \mathcal{B}_{min}^y}{N}\\
\end{aligned}
\end{equation}
The variables $\mathcal{B}_{max}$ and $\mathcal{B}_{min}$ respectively represent the maximum and minimum values of the scene's bounding box and $\ddot s^l$ is the spatial scale of level $l$ in the MipMap stack. In practice, we only store and optimize the features in level 0 MipMap $P_{x,y}^0$, while the remaining levels are dynamically computed and generated during forward inference. This way, we possess the capability to encode features at different spatial scales within the scene.

Meanwhile, for a Gaussian primitive with a scaling s in 4D space, when projected onto the spatial-only plane $P_{x,y}$, it results in a 2D ellipse with axes $(s_x,s_y)$.Therefore, based on the projected axes of the Gaussian primitive on $P_{x,y}$ and the corresponding base spatial scale ${\ddot s_{x,y}^0}$ of the MipMap stack, we can determine the spatial scale level associated with this Gaussian primitive:
\begin{equation}
    l^x = log2(\frac{ s_x}{\ddot s_x^{0}}) \quad\quad l^y= log2(\frac{ s_y}{\ddot s_y^{0}})
\end{equation}
To maintain the highest possible accuracy, we choose the minimum value among them as the final spatial level $l= min(l^x,l^y)$. So now we can obtain the two MipMap features that are closest to its spatial level:$P_{x,y}^{\lfloor l \rfloor},P_{x,y}^{\lceil l \rceil}$, and we can obtain the embedding of the Gaussian primitive with 4D position $\mu_{4d}$ in $P_{x,y}$ as:
\begin{equation}
    f_{x,y} = \psi_{tri}(\pi_{x,y}(\mu^{4d}),l;P_{x,y}^{\lfloor l \rfloor},P_{x,y}^{\lceil l \rceil},),
\end{equation}
Here, $\psi_{tri}$ represents trilinear interpolation in the space formed by $P_{x,y}^{\lfloor l \rfloor} $ and $ P_{x,y}^{\lceil l \rceil}$. The complete expression of the scale-aware residual feature $f$ of $\mathcal{G}^{4D}$ is as follows:
\begin{equation}
\begin{aligned}
% &f_{so} = \sum_{{i,j}\in C_{so}}Triinterp(\pi_{x,y}(\mu_{4D}),l;P_{x,y}^{\lfloor l \rfloor},P_{x,y}^{\lceil l \rceil},)\\
% &f_{st} = \sum_{{i,j}\in C_{st}} interp(\pi_{i,j}(\mu_{4D}) ; P_{i,j})\\
f(\mathcal{G}^{4D}) &= f_{so} + f_{st}\\
    &=\sum_{{i,j}\in C_{so}}\psi_{tri}(\pi_{x,y}(\mu^{4D}),l;P_{x,y}^{\lfloor l \rfloor},P_{x,y}^{\lceil l \rceil},)+\\
    &\quad\quad\sum_{{i,j}\in C_{st}} \psi_{bi}(\pi_{i,j}(\mu^{4D}) ; P_{i,j})\\
    C_{so} = &\{(x,y),(x,z),(y,z)\}\quad\quad C_{st} =\{(x,t),(y,t),(z,t)\}.
\end{aligned}
\label{eq:srhenc}
\end{equation}
Here, $\psi_{tri},\psi_{bi}$ represent  trilinear interpolation and bilinear interpolation respectively. Through experiments, we found that summation is a more effective way to combine features in our Scale-aware Residual Field compared to others.

\subsection{Adaptive Optimization}
\label{sec:optim}
%这里要提一下在4d空间中densify，并且会存在的问题
% Our optimization of Gaussian primitives is conducted directly in 4D space. In addition to optimizing the initial attributes of Gaussian primitive, we densify Gaussian primitives to optimize the currently imperfectly reconstructed regions. However, 
Due to the varying temporal  properties of Gaussian primitives in this 4D space, each Gaussian primitive is sampled with different probabilities over observed time. Dynamic primitives, in order to represent the temporal complexity of the scene, often have a smaller lifespan, resulting in a lower sampling probability compared to static primitives. 
% 在整个时域上曝光较少的点，会在损失函数反向传播过程中有较小的梯度. Therefore, applying the same optimization and densification strategy to each primitive with 3DGS\cite{3dgs} directly would 存在优化imbalance.
% The gradient value is 特别 important in 3DGS framework, because we need the gradient exceed the threhold to densify itself to optimize the currently imperfectly reconstructed regions. 
These dynamic primitives would have smaller gradients during the backward propagation of the loss function. 
The gradient value is crucial in the 3DGS framework, as it needs to exceed a threshold to densify the corresponding primitive and optimize the currently imperfectly reconstructed regions. Hence, applying the same optimization and densification strategy directly to each primitive with 3DGS\cite{3dgs} may lead to optimization imbalance.

To tackle the aforementioned issue, we propose an Adaptive Optimization strategy, which dynamically adjusts the learning rate and densify gradient threshold for $\mathcal{G}_i^{4D}$ based on its 
sampling probability across the observable time range.
% integration of its state function $\gamma_i(t)$ across the observable time range. 
Specifically, We can use $\gamma_i(t)$ of $\mathcal{G}_i^{4D}$ to calculate the temporal integral within the observable range, representing its sampling probability .
% This approach simultaneously considers both the temporal position and lifespan of the Gaussian. 
The larger the integral, the more the Gaussian primitive's lifespan intersects with the observable range, making it more likely to be sampled. 
% We propose a time-domain integration-based optimization schedule strategy, which dynamically adjusts the learning rate and densify gradient threshold for each Gaussian primitive based on the integration of its temporal lifespan opacity function across the observable time range.
% its probability of sampled over the observable range, i.e., the integration of its temporal lifespan opacity function across the time domain.

\noindent\textbf{Definite Integral of the Time Domain Distribution.}
Based on state function $\gamma_i(t)$ of $\mathcal{G}_i^{4D}$, we can compute its integral over the time domain.
\begin{equation}
  F(t) = P(x<t) = \int_{-\infty}^{t} e^{-k\frac{x-\tau_i}{\sigma_i}^2} dx
\end{equation}
\begin{equation}
  I = F(t_{end})-F(t_{start})
  \label{equl:intergral}
\end{equation}
$F(t)$ represents the CDF (cumulative distribution function) of each Gaussian primitive. Here, $I$ denotes the definite integral over the entire time domain from $t_{start}$ to $t_{end}$, where $t_{start}$ and $t_{end}$ normalize to 0 and 1, respectively.

Since $\gamma_i(t)$ is a Gaussian-like function, it is challenging to compute precise definite integral values. Inspired by \cite{intergralcount}, we derive the approximate cumulative distribution function as:
\begin{equation}
Q(t) = \int_{-\infty}^{t} \frac{1}{\sqrt{2\pi}} e^{-\frac{x^2}{2}}dx = 1- \frac{1}{e^{1+\alpha_1 t^3 + \alpha_2 t}}
  % F(t) = P(x<t) =  \frac{2\sqrt{\pi}s_{t}}{k}(1- \frac{1}{e^{1+\alpha_1 t^3 + \alpha_2 t}})
\end{equation}
\begin{equation}
    F(t) = \frac{\sqrt{\pi}\sigma_i}{\sqrt{k}} Q(\sqrt{2k}\frac{(t-\tau_i)}{\sigma_i})
\end{equation}
Here, $\alpha_1= 0.070565992,\alpha_2=1.5976$. Please refer to the appendix for details. Therefore, we can obtain the definite integral of Gaussian primitives over the time domain using Eq. \ref{equl:intergral} with minimal computational complexity.

\noindent\textbf{Intergral-based Per-Gaussian Optimization Schedule.}
After get the temoral intergral of each Gaussian, we can dynamically adjust the learning rates and gradient thresholds for densification control on a per-primitive basis, aiming to achieve rapid reconstruction of dynamic regions:
% To balance the optimization speeds between dynamic and static regions within the scene, we utilize the definite integral of Gaussian primitive's  temporal lifespan opacity function in the time domain to quantify its exposure across the observed time range. With this, we dynamically adjust the learning rates and gradient thresholds for densification control on a per-primitive basis, aiming to achieve rapid reconstruction of dynamic regions:

% \begin{equation}
% \setlength{\abovedisplayskip}{3pt}
% \setlength{\belowdisplayskip}{3pt}
%   \kappa_i= \kappa_{base}*\frac{I_i}{I_{max}}
% \end{equation}
\begin{equation}
\setlength{\abovedisplayskip}{3pt}
\setlength{\belowdisplayskip}{3pt}
  \kappa_i= \kappa_{base}*\frac{I_i}{I_{max}},
\quad\quad\quad
  lr_i= lr_{base}*\frac{I_{max}}{I_{i}}
  \label{eq:lr}
\end{equation}

% \textbf{Gradient Threshold schedule:}For the gradient threshold during density control, we scale the gradient threshold for densification based on the time-domain integral of each Gaussian primitive.
% \begin{equation}
%   \kappa_i= \kappa_{base}*\frac{I_i}{I_{max}}
% \end{equation}
Here, $\kappa_i$ and $lr_i$ respectively represent the densification threshold and the learning rate of $\mathcal{G}_i^{4D}$, while $I_i$ and $I_{max}$ respectively denote the timeporal integral of $\mathcal{G}_i^{4D}$ and the maximum time-domain integral among all Gaussian primitives. 
We adjust its densification threshold each time densification control is required. Additionally, for the learning rate, we dynamically adjust it every 50 iterations based on Eq. \ref{eq:lr}, involving parameters related to 4D position, scaling, rotation, and zeroth-order SH coefficients of $\mathcal{G}_i^{4D}$.

% 我们在每次要进行densification control时改变它的densification threshold。And 对于学习率，我们根据式21每50次迭代进行动态调整，涉及到的参数包括G的abx

% \textbf{Learning rate schedule:}For the learning rate of each Gaussian primitive, we dynamically adjust it every 50 iterations based on the time-domain integral of each primitive. The parameters involved for each Gaussian primitive include 4D position, scaling, rotation, and zeroth-order SH coefficients.
% \begin{equation}
%   lr_i= lr_{base}*\frac{I_{max}}{I_{i}}
% \end{equation}

% We also perform additional pruning based on the time-domain integral. Every 50 iterations, we filter out primitives with integrals smaller than a threshold to prevent excessive growth of these nearly invisible points during the entire observation time range.

\subsection{Loss Function}
\label{sec:loss}

\textbf{Regularization term for scaling residuals:}
% 当采样时间t等于G的时域position时，我们希望投影到3D空间中的Gaussian primitive的attributes相较于在4D空间中的初始值不要变化太大，否则投影过程会过度依赖于attributes的残差，从而忽略了对其初始值的优化.此外，我们希望Gaussian primitives在4D空间中的scaling的初始值与其在3D空间中的值的差距不要太大，这样momentum field才能结合更准确的尺度信息。为了实现这个目标，我们提出了一个关于scaling residuals的Regularization term：
When the sampling time 
$t$ equals the temporal position of $\mathcal{G}_i^{4D}$, we aim for minimal variations in the attributes of Gaussian primitives projected into 3D space compared to their initial values in 4D space. Excessive reliance on attribute residuals during the projection process may neglect the optimization of their initial values. Additionally, we strive to minimize the disparity between the initial scaling values of Gaussian primitives in 4D space and their values in 3D space. This ensures that the Scale-aware Residual Field can effectively integrate accurate scale information. To achieve this goal, we propose a regularization term $\mathcal{L}_{SR}$ concerning scaling residuals:
\begin{equation}
\setlength{\abovedisplayskip}{3pt}
\setlength{\belowdisplayskip}{3pt}
\mathcal{L}_{SR}(\mathcal{G}^{4D}) = \frac{1}{n}\sum_{i}||\Delta \mathbf{\Sigma}_i(\tau_i)[:3]||_2
\end{equation}
Here, 
$\Delta \mathbf{\Sigma}_i(\tau_i)[:3]$ represents the residuals of scaling of 
$\mathcal{G}_i^{4D}$ obtained during the projection process when the sampling time is 
$\mathcal{G}_i^{4D}$'s temporal position 
$\tau_i$, where 
$n$ is the total number of Gaussian primitives in the current 4D space.

\noindent\textbf{Total Loss Function}
Following \cite{3dgs}, we use the loss between the rendered image and the ground truth image, which includes an $\mathcal{L}_1$ term and a $\mathcal{L}_{D-SSIM}$ term. Combined with our regularization terms, the overall loss function is formulated as:
\begin{equation}
\setlength{\abovedisplayskip}{3pt}
\setlength{\belowdisplayskip}{3pt}
\mathcal{L} = (1-\lambda_1)\mathcal{L}_1 + \lambda_1\mathcal{L}_{D-SSIM} +\lambda_2 \mathcal{L}_{SR}
    \label{eq:loss}
\end{equation}
The settings for 
$\lambda_1$ and 
$\lambda_2$ can be referred to in Sec. \ref{sec:imple}.

\vspace{-2.5ex}

\section{IMPLEMENTATION DETAILS}
\label{sec:imple}

We implemented our work using the PyTorch\cite{pytorch} framework and open-source code based on 3DGS. We utilized the nvidiffrast\cite{nvdiffrast} library to compute the MipMap stack in the Scale-aware Residual Field, ensuring computational efficiency. We use Adam optimizer and retained certain implementations from 3DGS, including the fast differentiable rasterizer, hyperparameters, and opacity reset strategy. In the loss function Eq. \ref{eq:loss}, we set $\lambda_1$ to 0.2 and $\lambda_2$ to 0.8. We conducted training and testing using a single RTX 3090.

To enhance our rendering speed, we adopted a lossless baking strategy for the model during rendering. During rendering, we can pre-compute the features for each Gaussian primitive in the 4D space and for a given sampling time $t_0$, we filter out points where the survival status $\gamma(t_0)<0.001$. These filtered points indicate that they are inactive and invisible at time $t_0$, thereby reducing the computational overhead during MLP Decoder and Splatting. After testing, our baking strategy has enabled us to double our rendering speed. Please refer to the appendix for more details.

% To enhance our rendering speed, we adopted a lossless baking strategy for the model during rendering. Since the features extracted from the Scale-aware Residual Field are independent of the sampling time, we can pre-compute the features for each Gaussian primitive in the 4D space. Thus, during rendering, the conversion of Gaussian primitives from 4D space to 3D space incurs only the overhead of MLP inference. Additionally, during rendering, for a given sampling time $t_0$, we filter out points where the survival status $\gamma(t_0)<0.001$. These filtered points indicate that they are inactive and invisible at time $t_0)$, thereby reducing the computational overhead during MLP Decoder and Splatting. After testing, our baking strategy has enabled us to double our rendering speed.
% 失活的点。即我们根据式14对每个高斯点计算出a，如果a小于阈值，那么我们认为
\begin{table}
\caption{Quantitative results on the monocular synthesis dataset D-NeRF. FPS is measured at $400\times400$ . $\dagger$ denotes a dynamic Gaussian method. The evaluation of $^1$ is conducted at a resolution of 800, while the remaining methods are evaluated at 400}
\vspace{-10pt}
\resizebox{0.45\textwidth}{!}
{
\begin{tabular}{lcllcc}
    \toprule
Method                            & PSNR(dB)$\uparrow$                      & SSIM$\uparrow$ & LPIPS$\downarrow$ & FPS$\uparrow$    & Train Time  \\     \midrule
D-NeRF\cite{dnerf}        & 29.67     & 0.95 & 0.07  &0.06      &48h        \\
KPlanes-hybrid\cite{kplanes}     & 32.36         & 0.96 & -     & 0.97   & 52m              \\
TiNeuVox-B\cite{tineuvox}          & 32.67         & 0.97 & 0.04  & 1.5    & 28m         \\
V4D\cite{v4d}          & 33.72         & 0.98 & 0.02  & 2.08   & 6.9h                \\
HexPlane\cite{hexplane}     & 31.04 & 0.97 & 0.04  & 2.5    & 11m 30s             \\
\midrule
4DGS\cite{4dgssplatting}$\dagger^1$          & 34.05 & 0.98 & 0.02  & 82     & 20m              \\
4DGS-Realtime\cite{realtime4dgs}$\dagger$ & 34.09 & 0.98 & -     & -      & -                   \\
Ours          & 36.13 & 0.98 & 0.01 & 182.29 &   45m                  \\   
\bottomrule
\end{tabular}
}
\label{tab:dnerf}
% \vspace{5pt}
\vspace{-0.3cm}
\end{table}

\begin{table}
% \caption{1}
\vspace{-0.15cm}
\caption{Quantitative results on the monocular synthesis dataset D-NeRF at more resolutions.}
\vspace{-0.3cm}
\resizebox{0.43\textwidth}{!}{\begin{tabular}{lcllcc}
    \toprule
Method   & PSNR(dB)$\uparrow$ &FPS$\uparrow$ &Method   & PSNR(dB)$\uparrow$ &FPS$\uparrow$   \\ \midrule

4DGS(400x400) & 35.05 & 107
&4DGS(800x800) & 34.05 & 82\\
\textbf{Ours(400x400)} & \textbf{36.13} & \textbf{182}                
&\textbf{Ours(800x800)} & \textbf{35.24} & \textbf{149}                   \\
\bottomrule
\end{tabular}}
\label{tab:dnerf_comp}
% \vspace{5pt}
\vspace{-0.3cm}
\end{table}

\begin{figure}[]
  \centering
  \setlength{\abovecaptionskip}{0.cm}

  \includegraphics[width=0.7\linewidth]{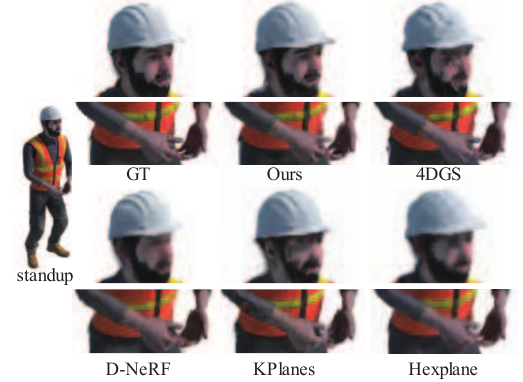}

  \caption{
Qualitative result on the D-NeRF dataset.}
\label{fig:dnerf}
\vspace{-0.6cm}
\end{figure}

\section{EXPERIMENTS}
% In this chapter, we initially introduce the single and multi-view datasets employed for evaluation in Sec. \ref{sec:dataset}. Subsequently, we present our evaluation results and comparisons with current state-of-the-art methods in Sec. \ref{sec:results}. Furthermore, we conduct ablation experiments in Sec. \ref{sec:ablation} to validate the effectiveness of the modules included in our approach. Finally, we discuss our limitations in Sec. \ref{sec:limition}
\subsection{Datasets}
\label{sec:dataset}
\textbf{Synthetic Dataset.}
We chose D-NeRF\cite{dnerf} as our evaluation dataset for monocular scenes. D-NeRF is a monocular synthetic dataset consisting of eight scenes with large-scale movements and real non-Lambertian material dynamic objects, which imposes a challenge on model performance. 

\noindent\textbf{Real-world Datasets.}
We selected Plenoptic Video dataset\cite{n3dLi} to evaluate our performance in multi-view real dynamic scenes. Plenoptic Video dataset consists of six real-world scenes, each captured by 15-20 cameras. Each scene in the dataset encompasses complex movements and occurrences of object appearance and disappearance. This dataset allows for a comprehensive evaluation of the model's reconstruction capability in complex temporal scenes.

% \vspace{-0.6cm}
\subsection{Results}
\label{sec:results}
% We chose D-NeRF\cite{dnerf} as our evaluation dataset for monocular scenes. D-NeRF is a synthetic dataset consisting of eight scenes with large-scale movements and real non-Lambertian material dynamic objects. Each scene contains 100-200 frames of views, with pre-defined divisions into training and testing sets. D-NeRF serves as a monocular dynamic dataset, meaning that in the training set, only one viewpoint image is available at any given moment, which imposes a challenge on model performance.
We have employed a variety of evaluation metrics to assess our model. For rendering quality, we utilize PSNR, SSIM, DSSIM, and LPIPS, and for rendering speed, we measure FPS. All evaluation results are averaged across all scenes in the dataset.

% We follow the division of training, validation, and test sets as in previous works. We employ PSNR ,SSIM and LPIPS as our evaluation metrics and
For monocular scenes in the D-NeRF dataset, we compare our method against the current state-of-the-art methods\cite{dnerf,kplanes,tineuvox,v4d,hexplane,4dgssplatting,realtime4dgs} in the field. The quantitative evaluation results are listed in Tab. \ref{tab:dnerf},\ref{tab:dnerf_comp}. The dynamic scene representation method based on NeRF\cite{dnerf,kplanes,tineuvox,v4d,hexplane} struggles with achieving real-time rendering due to the need for dense ray sampling during rendering. However, our method achieves a rendering speed of 182 FPS and exhibits considerable improvement in rendering quality. Compared to existing dynamic Gaussian methods\cite{4dgssplatting,realtime4dgs}, our approach demonstrates superior performance in handling complex temporal scenes and achieves a certain level of enhancement in rendering quality. 
Qualitative comparison results can be seen in Fig. \ref{fig:dnerf}. In the standup scene, our method demonstrates better reconstruction of details (such as facial and hand features).
% \subsection{Results on the multi-view real scenes}
% We selected Plenoptic Video dataset to evaluate our performance in multi-view real dynamic scenes. Plenoptic Video dataset consists of six real-world scenes, each captured by 15-20 cameras. We chose the first camera as the test viewpoint and the remaining cameras as the training viewpoints, consistent with previous work. Each scene in Neu3D has a duration of 10 seconds, totaling 300 frames, and encompasses complex movements and occurrences of object appearance and disappearance. This dataset allows for a comprehensive evaluation of the model's reconstruction capability in complex temporal scenes. 

\begin{table}[]
\setlength{\abovecaptionskip}{0.cm}

\caption{\textbf{Quantitative results on Plenoptic Video dataset.} FPS is measured at $1352\times1014$. $\dagger$ denotes a dynamic Gaussian method.$^1$: excludes the Coffee Martini scene. $^2$: Only report SSIM instead of MS-SSIM like others.}
\resizebox{0.45\textwidth}{!}{
\begin{tabular}{lcllcc}
    \toprule
Method   & PSNR(dB)$\uparrow$       & DSSIM$\downarrow$ & LPIPS$\downarrow$  & FPS\\    
\midrule
KPlanes-hybrid\cite{kplanes}     & 31.63         & - & -       &0.3            \\
Mix-Voxels-L\cite{mixvoxel}          & 31.34        & 0.017 & 0.096     &16.7    \\
NeRFPlayer\cite{nerfplayer}          & 30.69          &0.034$^2$  & 0.111    &0.045         \\
HyperReel\cite{hyperreel}     & 31.1  &0.037$^2$ & 0.096      &2.00            \\
StreamRF\cite{streamrf}     & 31.04  &- & 0.040           &8.3       \\

HexPlane\cite{hexplane}$^1$     & 31.71  &0.014 & 0.075      &-             \\
\midrule

4DGS-Realtime\cite{realtime4dgs}$\dagger$ & 32.01  & 0.014 & -         &114              \\
Spacetime-Gs\cite{spacetimegs}$\dagger$ & 32.05  & 0.014  & 0.044       &140              \\
4DGS\cite{4dgssplatting}$\dagger$          & 31.15  & 0.016 & 0.049      &30         \\
Ours          & 32.15  &0.014 & 0.044  &40                \\   
\bottomrule
\end{tabular}
}
\label{tab:n3d}
\vspace{-0.6cm}
\end{table}

\begin{figure*}[h]
  \centering
  \setlength{\abovecaptionskip}{0.cm}

    % \fbox{\rule{0pt}{2.5in} \rule{0.9\linewidth}{0pt}}
  \includegraphics[width=0.8\linewidth]{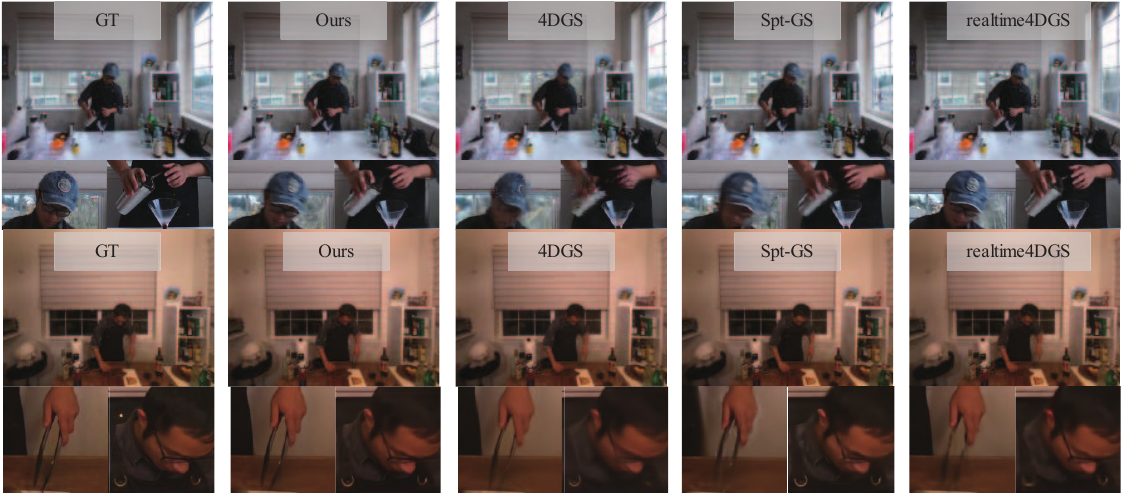}
  \caption{Qualitative results on coffee martinis and cut roasted beef from the Plenoptic Video dataset}
\label{fig:n3d}
\vspace{-0.6cm}

\end{figure*}
% We selected PSNR, DSSIM, and LPIPS as the evaluation metrics for assessing the reconstruction quality in real-world scenarios, and FPS as the benchmark for rendering speed.

For multi-view scenes in the Plenoptic Video dataset, our evaluation results are presented in Tab. \ref{tab:n3d}. As mentioned in \cite{spacetimegs,realtime4dgs}, there are two different calculation settings for DSSIM in previous works, and we have indicated this in the table. Our method outperforms all compared methods in rendering quality while maintaining superior rendering speed, achieving real-time rendering. The NeRF-based methods\cite{kplanes,mixvoxel,nerfplayer,hyperreel,streamrf,hexplane} have a significant disadvantage in rendering speed compared to ours. 4DGS\cite{4dgssplatting} based on deformation fields has shortcomings in modeling complex temporal situations such as object appearance and disappearance, which we can effectively address. Although Spacetime-GS\cite{spacetimegs} achieves higher rendering speeds, it is only applicable to multi-view scenes, whereas our method is suitable for both multi-view and single-view scenarios. 
A qualitative comparison of rendering quality with Gaussian methods can be seen in Fig. \ref{fig:n3d}. Our approach demonstrates more accurate reconstructions in temporally complex scenes (such as the inverted coffee scene with details on the heads and hands) and richer details (such as the circular decorations on the upper garment and reflections on the clippers in the cut beef scene).

\vspace{-0.3cm}
\subsection{Ablation Study}
% 我们对我们提出的和采用的方法进行了逐一的ablation study，定量的评测结果可见 Tab1，定性的结果可见Fig1. 在进行可视化展示时，we use a black background for training and test with a white
% background to better illustrate the impact of different strategies on SaRO’s ability to handle spatiotemporal information.
% We performed a thorough ablation study on our proposed and adopted methods, presenting quantitative results in Table 1 and qualitative results in Figure 1. Visual representation utilized a black background for training and a white background for testing, illustrating the impact of different strategies on SaRO's spatiotemporal processing capabilities.
\begin{figure}[]

  \centering
  \setlength{\abovecaptionskip}{0.2cm}

    % \fbox{\rule{0pt}{2.5in} \rule{0.9\linewidth}{0pt}}
  \includegraphics[width=0.8\linewidth]{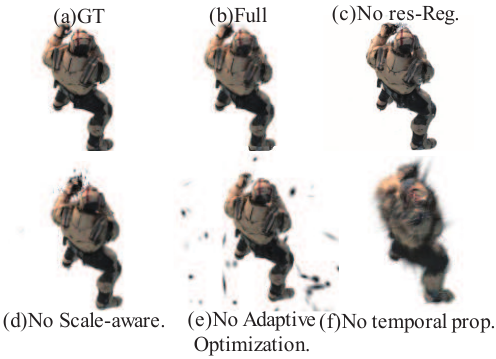}
  \vspace{-5pt}

  \caption{Qualitative results of the ablation study.
% During training, we use a black background and test with a white background to better illustrate the impact of different strategies on SaRO's ability to handle spatiotemporal information. 
}
\label{fig:ablation}
\vspace{-1.cm}
\end{figure}
\label{sec:ablation}

\begin{table}[]
\caption{The ablation study results across all scenes in the D-NeRF dataset.
% We conduct ablation study on our method across all scenes in the D-NeRF dataset.. 
% "No temporal prop." indicates that the temporal position and lifespan of the Gaussian are disregarded. "No res-Reg" indicates the absence of regularization for scaling residuals.
}
\vspace{-10pt}

\resizebox{0.37\textwidth}{!}{\begin{tabular}{lcllcc}
    \toprule
Method   & PSNR(dB)$\uparrow$       & SSIM$\uparrow$ \\    
\midrule
No Scale-aware & 35.61  & 0.98             \\
No temporal prop.    & 32.29  & 0.96          \\
No Adaptive Optimization & 35.44  & 0.98             \\
No res-Reg. & 35.43  & 0.98               \\
Full          & 36.13  &0.98            \\   
\bottomrule
\end{tabular}}
\label{tab:ablation}
\vspace{-20pt}

\end{table}
\noindent\textbf{Temporal properties of 4D Gaussian.}
% 在4D空间中的每个Gaussian primitive都有一个temporal position和一个lifespan，并且这个temporal position是分布在整个时域范围内的。这一点让我们在处理复杂的时域信息如物体的出现和消失的情况时优于以往的基于变形场的方法.为了验证这一点，我们让每个点的temporal position固定为0，即frame 0 的采样时间，and 我们采用Gaussian primitive的3D position和采样时间结合去momentum field中取特征，并且让Gaussian primitive在整个采样时间内去移动。结果如Tab. 3所示，没有了temporal position和lifespan，我们方法的性能会大幅下降。
% In 4D space, each Gaussian primitive has a temporal position and a lifespan, distributed across the entire temporal domain. This aspect
% 4D空间中的每个Gaussian均匀分布在整个时域上，并都有一个lifespan，
% gives us an advantage over previous deformation-based methods when dealing with complex temporal information. To validate this point, we fix the temporal position of each point to frame 0's sampling time and combine the 3D position of Gaussian primitives with the sampling time to extract features in the momentum field. Additionally, we allow Gaussian primitives to move throughout the entire sampling time. The results, as shown in Tab. \ref{tab:ablation}, indicate a significant performance degradation when temporal position and lifespan are removed.
Each Gaussian in 4D space possesses temporal properties, including temporal position and lifespan, giving us an advantage over previous deformation-based methods in handling intricate temporal information. To demonstrate this, we fix the temporal position of each point to frame 0 and 
enable Gaussians to traverse the entire sampling time. 
The results, as presented in Tab. \ref{tab:ablation} and Fig. \ref{fig:ablation}, show a notable decline in performance when temporal position and lifespan are disregarded.

\noindent\textbf{Consideration of Spatial Scale in the Residual Field.} In our Residual field, we take into account the size information of each Gaussian, improving the accuracy of the residual features. To evaluate this approach, we refrain from encoding the projection area in spatial-only planes. As depicted in Tab. \ref{tab:ablation} and Fig. \ref{fig:ablation}, not incorporating the size information of Gaussian primitives results in a reduction in the reconstruction quality of dynamic scenes.
% In our momentum field, we take into account the size information of each Gaussian, enhancing the accuracy of the features obtained for each Gaussian and consistent with the splitting of Gaussian primitives. To assess the effectiveness of this approach, we 不在spatio-only planes对投影区域编码， and conduct evaluations on the D-NeRF dataset. As shown in the Tab. \ref{tab:ablation}, when disregarding the size information of Gaussian primitives, our method exhibits a decrease in the reconstruction quality of dynamic scenes.
% exclusively utilize the 4D location of Gaussian primitives for feature projection and conduct evaluations on the D-NeRF dataset. As shown in the Tab. \ref{tab:ablation}, when disregarding the size information of Gaussian primitives, our method exhibits a decrease in the reconstruction quality of dynamic scenes.
% \textbf{特征之间通过加法相结合}

\noindent\textbf{Adaptive Optimization.} 
To address the imbalance in optimization between dynamic and static regions in the scene, we introduce the Adaptive Optimization strategy. Without this strategy, the reconstruction ability of 4D Gaussians for moving regions would decrease. 
This is supported by Tab. \ref{tab:ablation}. Without this strategy, Gaussian primitives struggle to distinguish between dynamic and static regions, leading to artifacts as shown in Fig. \ref{fig:ablation}(e).
% To address the issue of optimization imbalance caused by varying sampling probabilities of Gaussian primitives in 4D space over observation time, we introduce the Per-Gaussian optimization schedule strategy. By integrating Gaussian primitives over the temporal dimension to represent their sampling probabilities, different optimization strategies are set accordingly. Without this strategy, the reconstruction capability of 4D Gaussians for motion regions would decline. Tab. \ref{tab:ablation} corroborates this point, and visual results can be found in Figure 1.

\noindent \textbf{Regularization of scaling residuals.} We conducted evaluations without regularization of scaling residuals to validate the effectiveness of this regularization term. 
% The results are presented in Table 2. 
Without this constraint, Gaussian primitives overly rely on the Scale-aware Residual field, neglecting optimization of their own initial attributes, resulting in a decrease in model performance, as shown in Tab. \ref{tab:ablation} and Fig. \ref{fig:ablation}.

% \noindent \textbf{Baking}

% 这是之前基于变形场和动态Gaussian方法难以做到的。

% \noindent\textbf{Analysis of the number of Gaussian primitives in 4D space.} As Gaussian primitives are cloned or split, regions that are not yet well constructed will be optimized\cite{3dgs}, so it is meaningful to observe the number of Gaussian primitives and the corresponding reconstruction quality as the number of iterations increases.

% \noindent\textbf{lossless baking strategy} 
% 关于积分的更广泛的应用
% 点数的增长变化
% -随着训练变化
% -pertimestamp
% \vspace{-0.3cm}
\subsection{Limitations}
\label{sec:limition}
Our method achieves high-quality, real-time reconstruction of temporally complex scenes. However, it has limitations, the use of explicit and implicit mixing, such as 4D space plane decomposition and MLP inference, may reduce training speed.  Furthermore, there are many possible extensions to explore based on SaRO-GS, such as 4D content generation\cite{yin20234dgen}, dynamic object tracking\cite{dynamic3dgs}, and combining with Large Language Models\cite{achiam2023gpt} or Spoken Language Understanding\cite{serdyuk2018towards,cheng2023ml,cheng2023mrrl} to achieve more flexible interaction modes.
% The \verb|authornote| and \verb|authornotemark| commands allow a note
% to apply to multiple authors --- for example, if the first two authors
% of an article contributed equally to the work.

% If your author list is lengthy, you must define a shortened version of
% the list of authors to be used in the page headers, to prevent
% overlapping text. The following command should be placed just after
% the last \verb|\author{}| definition:
% \begin{verbatim}
%   \renewcommand{\shortauthors}{McCartney, et al.}
% \end{verbatim}
% Omitting this command will force the use of a concatenated list of all
% of the authors' names, which may result in overlapping text in the
% page headers.

% The article template's documentation, available at
% \url{https://www.acm.org/publications/proceedings-template}, has a
% complete explanation of these commands and tips for their effective
% use.

% Note that authors' addresses are mandatory for journal articles.
  \vspace{-0.3cm}
\begin{figure}[]
  \centering
  \setlength{\abovecaptionskip}{0.cm}

  \includegraphics[width=0.9\linewidth]{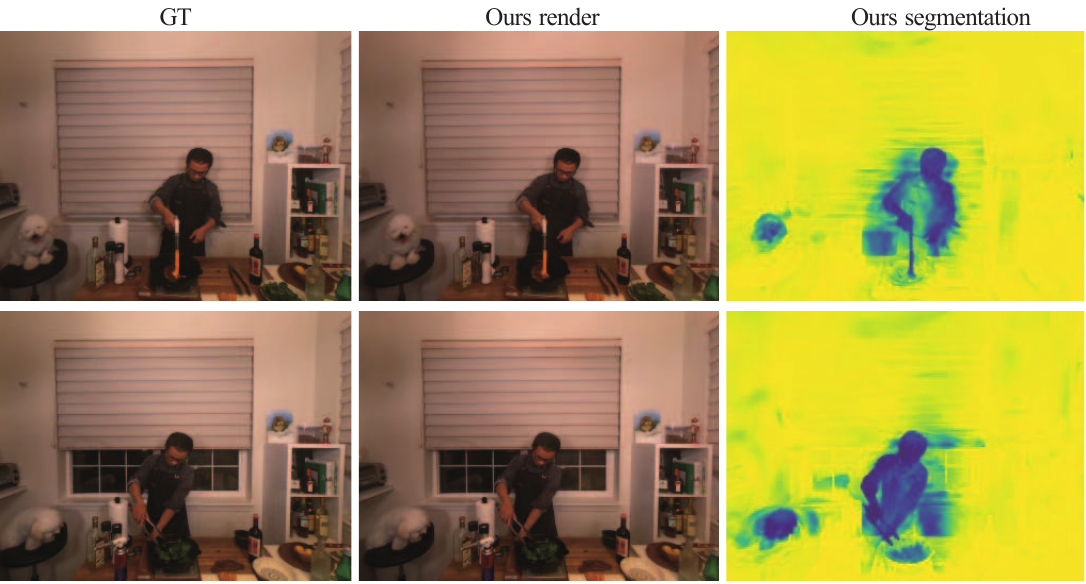}
  \caption{Segmentation of dynamic and static scenes. }
  \label{fig:segmentation}
  % \vspace{-0.8cm}
\end{figure}
\subsection{Dynamic-Static segmentation}
\label{sec:discussion}
\vspace{-0.1cm}

% \textbf{Dynamic-static scene segmentation.} 
Static primitives are observable throughout the entire observation period, whereas dynamic primitives are only visible near their temporal positions. Therefore, dynamic primitives have a shorter lifespan, while static primitives have a much longer lifespan. Consequently, we can segment the scene into dynamic and static parts based solely on their lifespans, as illustrated in Fig. \ref{fig:segmentation}. 
% Specifically, we use the lifespan of each Gaussian as a feature for Splatting, followed by mapping the resulting grayscale image to RGB space for visualization. 
Our method accurately segments dynamic foreground elements and dynamic lighting effects (e.g., shadows of people in the image).
This aspect also underscores the interpretability of our method.

\vspace{-0.3cm}
\section{CONCLUTION}

In this paper, we propose SaRO-GS as a novel approach for representing dynamic scenes, enabling real-time rendering while ensuring high-quality reconstruction, especially in temporally complex scenes. SaRO-GS utilizes a set of 4D Gaussians to represent dynamic scenes and leverages 3D Gaussian Splatting for real-time rendering. Additionally, we propose a Scale-aware Residual Field to encode the region occupied by Gaussian primitives, resulting in more accurate features and align with the self-splitting behavior of Gaussians. Furthermore, we introduce an Adaptive Optimization strategy to enhance the model's ability to reconstruct high-frequency temporal information in dynamic scenes. Experimental results in both monocular and multi-view settings demonstrate that SaRO-GS achieves SOTA rendering quality while enabling real-time rendering.

\begin{acks}
This work is financially supported by Guangdong Provincial Key Laboratory of Ultra High Definition Immersive Media Technology, this work is also financially supported for Outstanding Talents Training Fund in Shenzhen, Shenzhen Science and Technology Program-Shenzhen Cultivation of Excellent Scientific and Technological Innovation Talents project(Grant No. RCJC20200714114435057) , R24115SG MIGU-PKU META VISION TECHNOLOGY INNOVATION LAB.
\end{acks}
\bibliographystyle{ACM-Reference-Format}

\bibliography{sample-base}

%%
%% If your work has an appendix, this is the place to put it.

\renewcommand{\thesection}{\Alph{section}}
\renewcommand\thefigure{\Alph{section}\arabic{figure}}
\renewcommand\thetable{\Alph{section}\arabic{table}}
\setcounter{page}{1}
\setcounter{section}{0}
\setcounter{figure}{0}
\setcounter{table}{0}
\section{Overview}
With in the supplemtantary, we provide:
\begin{itemize}
    \item Details of Adaptive Optimization in Sec. \ref{sec:adaptive}
    \item Hyperparameter Settings in Sec. \ref{sec:hyper}
    \item More Results in Sec. \ref{sec:res}
    
\end{itemize}
% ACM's consolidated article template, introduced in 2017, provides a
% consistent \LaTeX\ style for use across ACM publications, and
% incorporates accessibility and metadata-extraction functionality
% necessary for future Digital Library endeavors. Numerous ACM and
% SIG-specific \LaTeX\ templates have been examined, and their unique
% features incorporated into this single new template.

% If you are new to publishing with ACM, this document is a valuable
% guide to the process of preparing your work for publication. If you
% have published with ACM before, this document provides insight and
% instruction into more recent changes to the article template.

% The ``\verb|acmart|'' document class can be used to prepare articles
% for any ACM publication --- conference or journal, and for any stage
% of publication, from review to final ``camera-ready'' copy, to the
% author's own version, with {\itshape very} few changes to the source.

\section{Details of Adaptive Optimization}
\label{sec:adaptive}
Based on the unique temporal characteristics of each Gaussian primitive, we apply distinct optimization schedules. Integrating the state function over time intervals enables us to represent the sampling probability of Gaussian primitives in the temporal domain: $I = F(t_{end})-F(t_{start})$, Here, $F(t)$ represents the cumulative distribution function (CDF) of the Gaussian primitive's state function $\gamma(t)$. 
\begin{equation}
  F(t) = P(x<t) = \int_{-\infty}^{t} e^{-k\frac{x-\tau}{\sigma}^2} dx,
\end{equation}
We approximate $F(t)$ based on\cite{intergralcount}:
% \begin{equation}
%   F(t) = P(x<t) = \int_{-\infty}^{t} e^{-k\frac{x-\tau}{\sigma}^2} dx
% \end{equation}
\begin{equation}
Q(t) = \int_{-\infty}^{t} \frac{1}{\sqrt{2\pi}} e^{-\frac{x^2}{2}}dx = 1- \frac{1}{e^{1+\alpha_1 t^3 + \alpha_2 t}},
  % F(t) = P(x<t) =  \frac{2\sqrt{\pi}s_{t}}{k}(1- \frac{1}{e^{1+\alpha_1 t^3 + \alpha_2 t}}),
\end{equation}

To transform it into the form of $F(t)$, we employ the method of integration by substitution:
\begin{equation}
x = \sqrt{2k} \frac{m-\tau}{\sigma}
\end{equation}

% \begin{equation}
% Q(t) = \int_{-\infty}^{\frac{\sigma t}{\sqrt{2k}}+\tau} \frac{1}{\sqrt{2\pi}} e^{-\frac{(\sqrt{2k} \frac{m-\tau}{\sigma})^2}{2}}d(\sqrt{2k} \frac{m-\tau}{\sigma})
% \end{equation}
% \begin{equation}
% = \int_{-\infty}^{\frac{\sigma t}{\sqrt{2k}}+\tau} \frac{1}{\sqrt{2\pi}} e^{-k\frac{m-\tau}{\sigma}^2} d(\sqrt{2k} \frac{m-\tau}{\sigma})
% \end{equation}
% \begin{equation}
% = \int_{-\infty}^{\frac{\sigma t}{\sqrt{2k}}+\tau} \frac{1}{\sqrt{2\pi}} e^{-k\frac{m-\tau}{\sigma}^2} d(\sqrt{2k} \frac{m-\tau}{\sigma})
% \end{equation}
% \begin{equation}
% = \int_{-\infty}^{\frac{\sigma t}{\sqrt{2k}}+\tau} \frac{1}{\sqrt{2\pi}} \frac{\sqrt{2k}}{\sigma} e^{-k\frac{m-\tau}{\sigma}^2} dm
% \end{equation}
% \begin{equation}
% = \frac{\sqrt{k}}{\sqrt{\pi}\sigma} \int_{-\infty}^{\frac{\sigma t}{\sqrt{2k}}+\tau}   e^{-k\frac{m-\tau}{\sigma}^2} dm
% \end{equation}
% \begin{equation}
% = \frac{\sqrt{k}}{\sqrt{\pi}\sigma} F(\frac{\sigma t}{\sqrt{2k}}+\tau)
% \end{equation}
\begin{align}
Q(t) =& \int_{-\infty}^{\frac{\sigma t}{\sqrt{2k}}+\tau} \frac{1}{\sqrt{2\pi}} e^{-\frac{(\sqrt{2k} \frac{m-\tau}{\sigma})^2}{2}}d(\sqrt{2k} \frac{m-\tau}{\sigma}) \\
=& \int_{-\infty}^{\frac{\sigma t}{\sqrt{2k}}+\tau} \frac{1}{\sqrt{2\pi}} e^{-k\frac{m-\tau}{\sigma}^2} d(\sqrt{2k} \frac{m-\tau}{\sigma}) \\
=& \int_{-\infty}^{\frac{\sigma t}{\sqrt{2k}}+\tau} \frac{1}{\sqrt{2\pi}} e^{-k\frac{m-\tau}{\sigma}^2} d(\sqrt{2k} \frac{m-\tau}{\sigma})\\
=& \int_{-\infty}^{\frac{\sigma t}{\sqrt{2k}}+\tau} \frac{1}{\sqrt{2\pi}} \frac{\sqrt{2k}}{\sigma} e^{-k\frac{m-\tau}{\sigma}^2} dm\\
=& \frac{\sqrt{k}}{\sqrt{\pi}\sigma} \int_{-\infty}^{\frac{\sigma t}{\sqrt{2k}}+\tau}   e^{-k\frac{m-\tau}{\sigma}^2} dm \\
    =& \frac{\sqrt{k}}{\sqrt{\pi}\sigma} F(\frac{\sigma t}{\sqrt{2k}}+\tau),
\end{align}
Therefore, we can obtain $F(t)$:
\begin{equation}
    F(t) = \frac{\sqrt{\pi}\sigma}{\sqrt{k}} Q(\sqrt{2k}\frac{(t-\tau)}{\sigma})
    \label{eq:f_t}.
\end{equation}

For each Gaussian primitive 
$\mathcal{G}_i^{4D}$
  with distinct $\sigma_i$ and 
$\tau_i$, we can derive 
$F_i(t)$ based on Eq. \ref{eq:f_t}. Then, we obtain $I_i$ for each 
$\mathcal{G}_i^{4D}$ , thus adopting different optimization schedules for each Gaussian primitive.

\section{HyperParameters settings}
\label{sec:hyper}
We predominantly adhere to the hyperparameter settings of 3DGS\cite{3dgs}. Specifically, the batch size in training is set to 4 , and we initialize the learning rate of Scale-aware Residual Field parameters at 3.2e-3, which decays to 3.2e-6 by the end of training. Similarly, we initialize the learning rates of all tiny MLP decoders at 1.6e-4, decaying to 1.6e-7 by the end of training. Furthermore, we opt to abandon the strategy of filtering out larger Gaussians in worldspace. Our decision stems from the observation that this strategy results in incomplete backgrounds in our framework, thereby compromising our rendering quality.

Different datasets are collected under different settings, corresponding to different initialization methods. For the D-NeRF dataset\cite{dnerf}, which involves monocular synthesized scene data, we uniformly and randomly initialize 10,000 Gaussian primitives distributed within a cube of $[-1.3, 1.3]^3$. Additionally, the temporal position of each Gaussian is uniformly initialized within the range of  $0$ to $1$. We adopt a warm-up strategy with 1,000 iterations to train the scene as static, compensating for geometric information loss due to the absence of initialized point clouds. 
We trained these synthesized scenes using a black background. Additionally, we set the opacity reset interval to 2,000 iterations to accelerate training. The total training duration is set to 20,000 iterations.

For the multi-view real-world Plenoptic Video dataset\cite{n3dLi}, we utilize point clouds generated by COLMAP as our initialization point cloud. To achieve more accurate scene boundaries, in addition to using point clouds from the first frame, we incorporate sparse point clouds generated from subsequent frames after undergoing sparse filtering. The total number of initial points for each scene is around 40,000. The temporal position of each Gaussian is also uniformly initialized within the range of 0 to 1. In this setting, warm-up is not necessary. We set
the opacity reset interval to 3,000 iterations  to align with 3DGS\cite{3dgs}.

\section{More Results}
Comparing with state-of-the-art (SOTA) methods\cite{dnerf,kplanes,tineuvox,v4d,hexplane,4dgssplatting}, the per-scene evaluation results are presented in Tab. \ref{tab:dnerf} for the D-NeRF dataset. Similarly, for the Plenoptic Video dataset, the per-scene evaluation results compared with SOTA methods\cite{kplanes,mixvoxel,nerfplayer,hyperreel,hexplane,dynamic3dgs,4dgssplatting,realtime4dgs,spacetimegs} are shown in Tab. \ref{tab:n3d}.

\begin{figure*}[]
  \centering
  % \setlength{\abovecaptionskip}{0.cm}

    % \fbox{\rule{0pt}{2.5in} \rule{0.9\linewidth}{0pt}}
  \includegraphics[width=\linewidth]{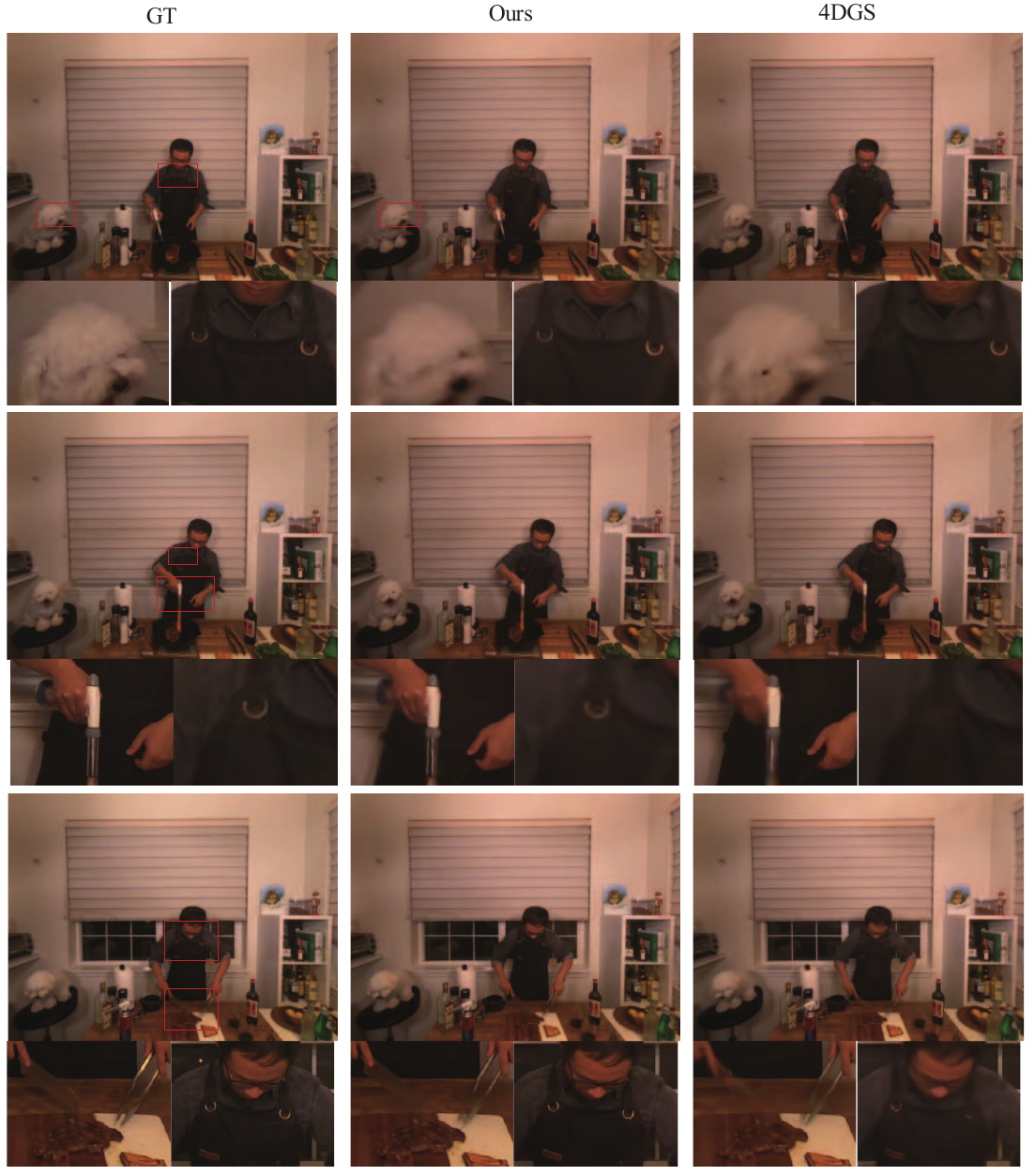}

  \caption{Qualitative comparison with \cite{4dgssplatting} on the Plenoptic Video dataset. }
\label{fig:appendix_n3d}
\end{figure*}

\begin{figure*}[]
  \centering
  % \setlength{\abovecaptionskip}{0.cm}

    % \fbox{\rule{0pt}{2.5in} \rule{0.9\linewidth}{0pt}}
  \includegraphics[width=\linewidth]{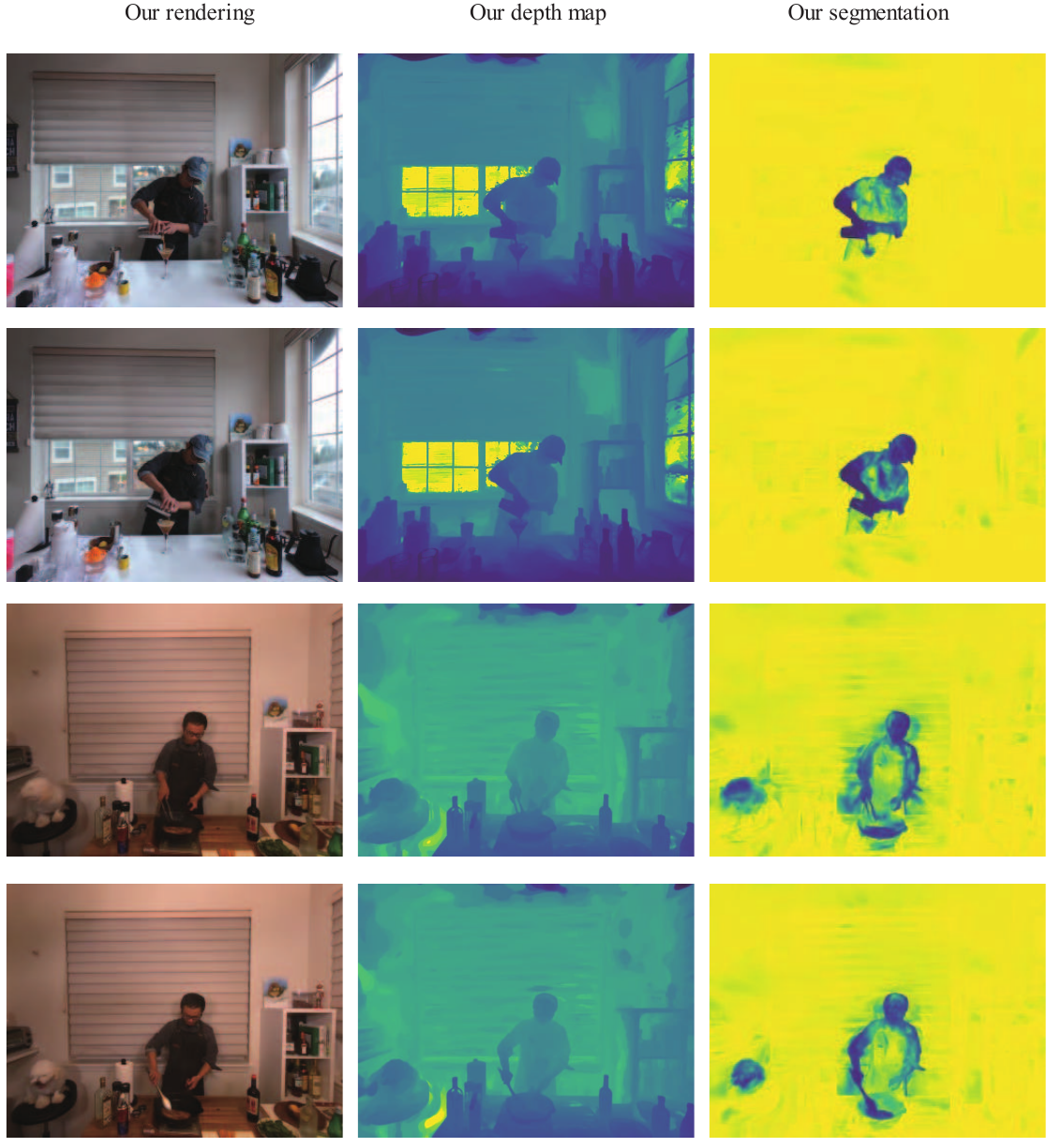}

  \caption{Depth map and  Segmentation of dynamic and static scenes.}
\label{fig:depth_seg}
\end{figure*}

In comparison with \cite{4dgssplatting}, more qualitative comparisons are presented in Fig. \ref{fig:appendix_n3d}. 
More results regarding dynamic-static segmentation in real-world scenes are presented in Fig. \ref{fig:depth_seg}. Additionally, the depth map  can alse be obtained through Gaussian splatting during the rendering process,as shown in Fig. \ref{fig:depth_seg}.

\begin{table*}
\caption{The per-scene evaluation results on the D-NeRF dataset.$\dagger$ denotes a dynamic Gaussian method.}
\resizebox{\textwidth}{!}{\begin{tabular}{lcllcllcllcll}
    \toprule
    
\multirow{2}{*}{Method}      & \multicolumn{3}{c}{Bouncing Balls}   & \multicolumn{3}{c}{Hellwarrior} & \multicolumn{3}{c}{Hook} & \multicolumn{3}{c}{Jumpingjacks}    \\ 
% \cline{2-4}   \cline{5-7} \cline{8-10} \cline{11-13}
\cmidrule(r){2-4}  \cmidrule(r){5-7} \cmidrule(r){8-10} \cmidrule(r){11-13}
 ~  &PSNR(dB)$\uparrow$ & SSIM$\uparrow$ & LPIPS$\downarrow$ &PSNR(dB)$\uparrow$ & SSIM$\uparrow$ & LPIPS$\downarrow$ &PSNR(dB)$\uparrow$ & SSIM$\uparrow$ & LPIPS$\downarrow$ &PSNR(dB)$\uparrow$ & SSIM$\uparrow$ & LPIPS$\downarrow$\\\midrule
D-NeRF\cite{dnerf} &38.93 &0.98 &0.10 &25.02 &0.95 &0.06  &29.25 &0.96 &0.11 &32.80 &0.98 &0.03  \\ 
KPlanes-hybrid\cite{kplanes}    &40.33 &0.99 &- &24.81 &0.95 &- &28.13 &0.95 &- &31.64 &0.97 &-   \\
TiNeuVox-B\cite{tineuvox}   &40.73 &0.99 &0.04  &28.17 &0.97 &0.07 &31.45 &0.97 &0.05 &34.23 &0.98 &0.03    \\
V4D\cite{v4d}  &42.67 &0.99 &0.02 &27.03 &0.96 &0.05 &31.04 &0.97 &0.03 &35.36 &0.99 &0.02\\
HexPlane\cite{hexplane}    &39.69 &0.99 &0.03 &24.24 &0.94 &0.07 &28.71 &0.96 &0.05 &31.65 &0.97 &0.04  \\
\midrule
4DGS\cite{4dgssplatting}$\dagger$   &40.62 &0.99 &0.02 &28.71 &0.97 &0.04 &32.73 &0.98 &0.03 &35.42 &0.99 &0.01      \\
% 4DGS-Realtime\cite{realtime4dgs}\dagger & 34.09 & 0.98 & -    &&& &&& &&&                   \\
Ours          & 36.02 & 0.99 &0.01 &38.01&0.97&0.02   &36.81&0.99&0.01     &34.56&0.98&0.016       \\   
\bottomrule

\multirow{2}{*}{Method}      & \multicolumn{3}{c}{Lego}   & \multicolumn{3}{c}{Mutant} & \multicolumn{3}{c}{Standup} & \multicolumn{3}{c}{Trex}    \\ 
% \cline{2-4}   \cline{5-7} \cline{8-10} \cline{11-13}
\cmidrule(r){2-4}  \cmidrule(r){5-7} \cmidrule(r){8-10} \cmidrule(r){11-13}
 ~  &PSNR(dB)$\uparrow$ & SSIM$\uparrow$ & LPIPS$\downarrow$ &PSNR(dB)$\uparrow$ & SSIM$\uparrow$ & LPIPS$\downarrow$ &PSNR(dB)$\uparrow$ & SSIM$\uparrow$ & LPIPS$\downarrow$ &PSNR(dB)$\uparrow$ & SSIM$\uparrow$ & LPIPS$\downarrow$\\\midrule
D-NeRF\cite{dnerf}        &21.64 &0.83 &0.16 &31.29 &0.97 &0.02  &32.79 &0.98 &0.02  &31.75 &0.97 &0.03  \\ 
KPlanes-hybrid\cite{kplanes}     &25.27 &0.94 &- &32.59 &0.97 &- &33.17 &0.98 &- &30.75 &0.97 &-      \\
TiNeuVox-B\cite{tineuvox}    &25.02 &0.92 &0.07  &33.61 &0.98 &0.03 &35.43 &0.99 &0.02  &32.70 &0.98 &0.03    \\
V4D\cite{v4d}        &25.62 &0.95 &0.04   &36.27 &0.99 &0.01  &37.20 &0.99 &0.01  &34.53 &0.99 &0.02    \\
HexPlane\cite{hexplane} &25.22 &0.94 &0.04 &33.79 &0.98 &0.03 &34.36 &0.98 &0.02 &30.67 &0.98 &0.03  \\
\midrule
4DGS\cite{4dgssplatting}$\dagger$       &25.03 &0.94 &0.04 &37.59 &0.99 &0.02 &38.11 &0.99 &0.01 &34.23 &0.99 &0.01          \\
% 4DGS-Realtime\cite{realtime4dgs}\dagger & 34.09 & 0.98 & -    &&& &&& &&&                   \\
Ours          & 25.46 & 0.94 & 0.04 &42.11&0.99&0.01   &44.45&0.99&0.01     &31.62&0.98&0.01         \\   
\bottomrule
\end{tabular}}
\label{tab:dnerf}
% \vspace{5pt}
% \vspace{-0.6cm}
\end{table*}

\begin{table*}
\caption{The per-scene evaluation results on the Plenoptic Video dataset.$\dagger$ denotes a dynamic Gaussian method.}
\resizebox{\textwidth}{!}{\begin{tabular}{lcllcllcllcll}
    \toprule
    
\multirow{2}{*}{Method}      & \multicolumn{3}{c}{Coffee Martini}   & \multicolumn{3}{c}{Cook Spinach} & \multicolumn{3}{c}{Cut Beef}    \\ 
\cmidrule(r){2-4}  \cmidrule(r){5-7} \cmidrule(r){8-10} 
 ~  &PSNR(dB)$\uparrow$ & DSSIM$\downarrow$ & LPIPS$\downarrow$ &PSNR(dB)$\uparrow$ & DSSIM$\downarrow$ & LPIPS$\downarrow$ &PSNR(dB)$\uparrow$ & DSSIM$\downarrow$ & LPIPS$\downarrow$ \\\midrule
KPlanes-hybrid\cite{kplanes}     &29.99 &0.024 &- &32.60 &0.017 &- &31.82 &0.017 &-      \\
Mix-Voxels-L\cite{mixvoxel}     &29.63 &0.024 &0.106 &32.25 &0.016 &0.099 &32.40 &0.016 &0.088      \\
NeRFPlayer\cite{nerfplayer}        &31.53 &- &0.085 &30.56 &- &0.113  &29.35 &- &0.144   \\ 
HyperReel\cite{hyperreel}    &28.37 &- &0.127  &32.30 &- &0.089 &32.92 &- &0.084     \\
% StreamRF\cite{streamrf}        &25.62 &0.95 &0.04   &36.27 &0.99 &0.01  &37.20 &0.99 &0.01    \\
HexPlane\cite{hexplane} &- &- &- &32.04 &0.015 &0.082 &32.55 &0.013 &0.080  \\
\midrule
Dynamic 3DGS\cite{dynamic3dgs}$\dagger$ &26.49 &0.033 &0.139 &32.97&0.013&0.087 &30.72 &0.016&0.090\\
4DGS-Realtime\cite{realtime4dgs}$\dagger$   &28.33 &- &- &32.93 &- &- &33.85 &- &-     \\
Spacetime-Gs\cite{spacetimegs}$\dagger$   &28.61 &0.025 &0.069 &33.18 &0.011 &0.037 &33.52 &0.011 &0.036     \\
4DGS\cite{4dgssplatting}$\dagger$         &27.34 &0.048 &- &32.46 &0.026 &- &32.90 &0.022 &-    \\
Ours          & 28.96 & 0.021 &0.061 &33.19&0.012&0.038   &33.91&0.021&0.038             \\   
\bottomrule

\multirow{2}{*}{Method}      & \multicolumn{3}{c}{Flame Salmon}   & \multicolumn{3}{c}{Flame Steak} & \multicolumn{3}{c}{Sear Steak}   \\ 
% \cline{2-4}   \cline{5-7} \cline{8-10} \cline{11-13}
\cmidrule(r){2-4}  \cmidrule(r){5-7} \cmidrule(r){8-10} 
 ~  &PSNR(dB)$\uparrow$ & DSSIM$\downarrow$ & LPIPS$\downarrow$ &PSNR(dB)$\uparrow$ & DSSIM$\downarrow$ & LPIPS$\downarrow$ &PSNR(dB)$\uparrow$ & DSSIM$\downarrow$ & LPIPS$\downarrow$ \\\midrule

KPlanes-hybrid\cite{kplanes}     &30.44 &0.024 &- &32.38 &0.015 &- &32.52 &0.013 &-       \\
Mix-Voxels-L\cite{mixvoxel}     &29.81 &0.026 &0.116 &31.83 &0.014 &0.088 &32.10 &0.012 &0.080      \\
NeRFPlayer\cite{nerfplayer}        &31.65 &- &0.098 &31.93 &- &0.088  &29.13 &- &0.138    \\ 
HyperReel\cite{hyperreel}    &25.02 &- &0.136  &33.61 &- &0.078 &35.43 &- &0.077      \\
% StreamRF\cite{streamrf}        &25.62 &0.95 &0.04   &36.27 &0.99 &0.01  &37.20 &0.99 &0.01     \\
HexPlane\cite{hexplane} &29.47 &0.018 &0.078 &31.82 &0.012 &0.071 &32.23 &0.012 &0.070   \\
\midrule
Dynamic 3DGS\cite{dynamic3dgs}$\dagger$ &26.92 &0.030 &0.122 &33.24&0.011&0.079 &33.68 &0.011&0.079\\
4DGS-Realtime\cite{realtime4dgs}$\dagger$   &29.38 &- &- &34.03 &- &- &33.51 &- &-       \\
Spacetime-Gs\cite{spacetimegs}$\dagger$  &29.48 &0.022 &0.063 &33.64 &0.009 &0.029 &33.89 &0.009 &0.030      \\
4DGS\cite{4dgssplatting}$\dagger$         &29.20 &0.042 &- &32.51 &0.023 &- &32.49 &0.022 &-          \\
% 4DGS-Realtime\cite{realtime4dgs}\dagger & 34.09 & 0.98 & -    &&& &&& &&&                   \\
Ours          & 29.14 & 0.021 & 0.059 &33.83&0.010&0.034 &33.89&0.010&0.036             \\   
\bottomrule
\end{tabular}}
\label{tab:n3d}
% \vspace{5pt}
% \vspace{-0.6cm}
\end{table*}

\label{sec:res}

\end{document}